\documentclass[11pt]{article}

\usepackage[margin=1in]{geometry}
\usepackage[T1]{fontenc}
\usepackage[utf8]{inputenc}
\usepackage{graphicx}
\usepackage{booktabs}
\usepackage{amsmath}
\usepackage{caption}
\usepackage{listings}
\usepackage{xcolor}
\usepackage[hidelinks]{hyperref}

\graphicspath{{figures/}}

\newenvironment{references}
  {\section*{References}\begin{list}{}{\setlength{\leftmargin}{2em}\setlength{\itemindent}{-2em}\setlength{\itemsep}{4pt}\setlength{\parsep}{0pt}}}
  {\end{list}}

\title{\bfseries Travel Time Prediction in Supply Chain Management\\Using Machine Learning\thanks{This paper is based on the author's doctoral dissertation completed at the Swiss School of Business and Management, Geneva (2024).}}
\author{Balaji Venkateswaran}
\date{March 2024}

\begin{document}

\maketitle

\begin{abstract}
\noindent The purpose of this research is to find data and methods using machine learning and deep learning to correctly predict the estimated travel time for transportation and logistics in a supply chain system. The supply chain ecosystem is very complex and heavily relies on the transportation and logistics of raw materials and finished goods. Accurate travel time estimation is critical because it helps supply chain members to improve logistics consistency and performance. This helps in planning, demand forecasting, lead time management and assembly planning. The logistics on the delivery side of the customer also plays a crucial role in customer satisfaction and voice of customer. With the collection of huge historical data and using novel techniques, the research builds an accurate model to predict travel time of inventory.
\end{abstract}


\section{Introduction}

Given that it enables supply chain operators to improve the quality of their operations, an accurate trip time prediction is extremely valuable for freight transports. In order to maximise projections of material supplies, a material planner at the receiving facility can identify impending delays in deliveries. Additionally, a factory can boost efficiency by adjusting its capacity over time, such as employees or equipment. The same advantages apply to logistic service suppliers. It is possible to plan employees, ramps, forklifts, and other resources appropriately in warehouses, ports, and other hubs (Abdollahi et al., 2020).

As a consequence, manufacturers and logistic service providers can improve their productivity, streamline their workflows, and boost planning precision. Generally speaking, a supply chain is made up of all cooperative operations by all participating businesses to transform raw materials into the finished product. This covers tasks like locating raw supplies, producing goods, putting them all together, and distributing them to final consumers. Managing, moving, and storing resources logistically is necessary to do this. Unlike supply chain management, which focuses on the planning and administration of supply chains, logistics addresses the operational level (Shahbazi \& Byun, 2020).

Information sharing throughout the various supply chain stages is another responsibility included in logistics. Usually, a variety of businesses or organizations are in charge of transportation. These circumstances greatly increase the supply chain's complexity, particularly in multimodal transports with numerous transshipment sites. The precise estimation of trip time in multimodal transports is the subject of this work. For the effective administration of transportation operations and logistics in supply chains, accurate trip time estimations are crucial. Continuous monitoring of freight shipments is necessary for estimating journey times, for example, utilizing mobile sensors attached to the goods being transported (Fu et al., 2020).

However, one of the most difficult logistics responsibilities is achieving the necessary transparency. It can be challenging to anticipate travel times because so many variables, like the weather, traffic, vehicles, routes, and transit relationships, can affect it. The majority of the material that is now available focuses more on passenger transportation than freight transport, such as bus arrival times or highway trip times (Sharma et al., 2020).

This limits the number of rides that can be included to those that last up to two hours; freight transports are not included. Complex and non-linear interactions between predictors can be dealt with more effectively by ML, which is also capable of processing complex and noisy data. This assertion is supported by the literature, which also shows that ML techniques typically outperform average-based approaches in terms of performance. Although this is the case, ML has only been utilized in a small number of recent articles that deal with journey time prediction in freight shipments (Kantasa-ard et al., 2021).

In transportation systems, travel time is a benchmark measurement. Consequently, the advancement of advanced traffic management systems (ATMS), intelligent transportation systems (ITS), and other transportation systems depends greatly on its precise prediction. From a traveler's point of view, knowing how long it will take to get somewhere and how much traffic there is might assist them choose faster routes, calculate how long it will take to get there, and choose better roads (Liu et al., 2020).

A precise technique for predicting travel times would also help transportation organizations manage traffic and ease congestion more effectively. Traffic data can be used to assess and forecast trip times and can be gathered from a variety of sources, including video cameras, automatic number plate recognition (ANPR), cellular geolocation, global positioning systems (GPS), automatic vehicle identification systems (AVI), and more. According to the particular objective and task, these sources may be employed in ITS (Kamble et al., 2021).

Since each GPS may record some relevant information, including latitude, longitude, a time stamp, speed, and other metadata, it is generally agreed upon that GPS data is the best source for predicting journey times. Design and operation, transportation management and planning, measurements, and assessment can all benefit from an objective, low variance estimation of journey time. In ATIS, trip duration information is particularly valued as essential pre-trip or even en-route information. They greatly enlighten both drivers and passengers, as was already noted, allowing them to plan their journeys more effectively or make wiser decisions (Tang et al., 2020).

In general, there are three advantages to travel time estimation. From the standpoint of the traveler, route selection before and during the trip is made easier and more reliable with the aid of trip duration information. This information may be used in logistics applications to boost delivery dependability, decrease delivery costs, or even enhance service quality. Keep in mind that this data is a critical performance indicator for the operation of the traffic system for traffic planners and managers (Abdollahi et al., 2020).

Using a combination of machine learning and data mining approaches, predictive analytics is a subfield of data engineering that makes predictions about the future based on study of historical data or events. An understanding of the business domain, data, and analytical methods is necessary for predictive analytics applications. Predictive analysis has applications in a wide range of industries, including finance, telecommunications, insurance, customer service, and many more (Chen et al., 2020). The telecommunications and healthcare industries have made extensive use of descriptive analytical tools, such as process and data mining techniques, to enhance operations and offer better customer service.

Logistics and transportation are one area of the supply chain industry that can use this data to raise the production and caliber of their operations. Industries all over the world rely largely on the logistics and transport sector to maintain the continuity of their supply chain. The industry for air freight transportation has grown as a result of industries shifting over time in favor of this mode of transportation. This is a fantastic chance to use air transport data to get insightful knowledge and assist businesses that both directly and indirectly touch this industry in improving their business operations (Abbasi et al., 2020).

The logistics and transportation sector must transition to sophisticated data analytics methodologies, such as proactive and predictive analytics, to make choices quickly and effectively by utilizing the full potential of data analytics, in order to enhance corporate performance and ensure competitiveness. Infrastructure, networks, IT, and many stakeholders are just a few of the functions that are combined in the logistics and transportation sector. This makes the entire process incredibly difficult and raises problems with effectively exporting commodities around the world. Utilizing technology, logistic organizations can adapt to changing needs and meet rising demand (Chen et al., 2021).

Organizations in the transportation sector are attempting to optimize their operations by using cutting-edge data analytics tools to examine resources, boost operational effectiveness, and forecast demand. These businesses will be able to make better judgments with the aid of predictive analytics. The transportation and logistics sectors have experienced considerable growth in recent years, which has made it difficult for businesses to find safe and reliable transportation options. When managing their shipments, logistics companies, like air cargo companies, run into a number of problems (Wu et al., 2020).

Delays in supply chain shipments being delivered is one issue these businesses encounter. Due to the high expense of air transportation, any delay in a shipment causes an interruption in their supply chains and resulting in a large financial loss. Early anticipation of any shipping process delays is difficult due to the convoluted structure and high level of uncertainties. Time-sensitive and more valuable items that need to travel over longer distances are catered to by the air cargo sector (He et al., 2020).

Given that it enables supply chain players to improve the quality of their operations, an accurate trip time forecast is extremely valuable for freight transportation. In order to maximize projections of material supplies, a material organizer at the receiving facility can identify impending delays in delivery. A factory can also boost its efficiency by adjusting its capacity over time, such as its employees or equipment (Akbari \& Do, 2021).

Similar advantages accrue to providers of logistical services. The availability of personnel, ramps, forklifts, and other resources may be arranged appropriately at warehouse, port, or other hubs. As a result, manufacturer including logistic service providers may improve their productivity, streamline their operations, and boost planning precision. In general, a supply chain comprises all cooperative activities carried out by all participating businesses to turn raw materials towards the finished product. This involves tasks like obtaining raw materials, producing goods, putting them together, and distributing them to customers (Kong et al., 2021).

Logistics for such handling, delivery, and storage of goods are necessary to accomplish this. Unlike supply chain management, which focus on the planning and administration of supply networks, logistics deals well with operational level. Providing information between such various supply chain stages is a duty that falls under logistics. The transport is often handled by a variety of businesses or organizations. These circumstances greatly increase the complexity of the supply chain, specifically in multimodal transport with several transshipment sites. This research focuses on providing reliable trip time estimates in multimodal transportation (Helo \& Hao, 2021).

For the effective administration of transportation logistics and operations in supply chains, accurate trip time estimations are crucial. For the purpose of estimating trip time, it is necessary to continuously monitor freight shipments, maybe by utilizing mobile sensors connected to the cargo. At the same moment, among the most difficult logistical responsibilities is ensuring the necessary openness. It is challenging to anticipate travel times since so many variables, including weather, traffic, vehicles, routes, and transportation relationships, may affect them (Kantasa-ard et al., 2021).

For transportation, accurate travel time estimation is critical because it helps supply chain members to improve logistics consistency and performance. It necessitates adequate prediction methods as well as proper input data, which can be provided, for example, by mobile sensors (Videsh D., 2021). With the collection of IoT data, routing determined by network optimizations and different parameters available, some predictive analytics can be done for travel time to help the entire supply chain.

An accurate trip time forecast is extremely valuable for freight transportation, and it enables supply chain players to improve the quality of their operations. To maximize projections of material supplies, a material organizer at the receiving facility can identify impending delays in delivery.

In the proposed method, we thus want to ascertain if supply chain delay may be forecast by statistically evaluating historical delay utilizing deep algorithms. The threat of supply chain interruptions might be decreased by using our forecasts to the advantage of shipping participants like sender, carrier, terminus operator, as well as recipient. Therefore, by using a unique and effective technique like the one employed in this work, it is possible to simply anticipate the time transit in the supply chain by using an effective deep learning approach.

\subsection{Research Background and Scope}

Machine Learning (ML) algorithms are ideal for resolving non-linear and dynamic interactions in monitoring data (Carbonneau et al., 2008). For transports, a reliable travel time forecast is extremely valuable because it helps supply chain members to improve their logistics efficiency (Lei L., et al., 2019). An inventory manager at the receiving plant will predict postponed deliveries in advance, allowing for better material stock forecasting. A plant may also change its capacities, such as workers or equipment, over time to improve its performance. In the same way, logistic service providers profits. Staffing, bridges, forklifts, and other capacities may be scheduled accordingly at stores, ports, and other hubs (Servos et al., 2019).

Simultaneously, ensuring the requisite clarity is one of the most difficult logistical challenges. Predicting travel time is also difficult because several factors affect it, including temperature, traffic, and driver, direction, and transportation relationship (Servos et al., 2019).

\subsection{Research Problem}

Transportation and logistics form one of the core components of supply chain management. For transportations, accurate travel time estimation is critical because it helps supply chain members to improve logistics consistency and performance. The arrival time of each inventory raw material determines the planning for manufacturing and assembly. This in turn helps deliver the finished products or downstream raw materials on time. The logistics on the delivery side of the customer also plays a crucial role in customer satisfaction and voice of customer.

In supply chain management, shipment deliveries play a crucial role in global transportation of goods which connects production to customer. In particular, predicting the time of arrival of delivery of shipment is absolute necessary when it comes to supply chain. Various authors have analyzed ways in which Estimated time arrival (ETA) can be predicted. Most of them have used machine learning linear regression models (Servos et al., 2019; Viellechner \& Spinler, 2020) to predict the ETA; they have used linear models which have lesser power to predict the correct ETA when given a complex dataset. Performing modelling using advanced deep learning techniques like recurrent neural networks or long short-term memory can acquire a higher performance level for predicting ETA.

\subsection{Research Aims}

The research aim is to use machine learning advanced deep learning techniques to correctly predict the estimated travel time for transportation and logistics in a supply chain system.

The scope of the transportation is only shipments on waterways on cargo ships.

\subsection{Research Objectives}

The aim of this study is to develop an advanced deep learning (1D-CNN) technique to estimate the travel time of deliveries in supply chain management sector. The objective of the research can be explained through the following:
\begin{itemize}
  \item To collect the data through online sources (shipment details, weather data, geo-location data).
  \item To preprocess the data using data cleaning techniques, removing duplicates, removing null elements and data aggregation.
  \item The pre-processed data is then sent to the feature selection algorithms, where the features with higher importance is selected either using principal component analysis or random forest regressor.
  \item One dimensional convolutional neural network (1D-CNN) is then applied on the selected features to produce the feature maps which contain important information from the data.
\end{itemize}

\subsection{Purpose of Research}

The purpose of this research is to have an accurate ETA determination model based on historical data to determine travel times for future transportation of shipments.

\subsection{Significance of the Study}

The study is significant because it demonstrates the variety of supply chain interruptions that may be caused, mostly by extreme weather events including traffic congestion. Travel time prediction would be a hard issue with supply chain management since probable traffic states and traffic occurrences also fall under the category of hidden modes. It is crucial to do research to enhance the transport service in such cities since such a large population relies on mass transit. Due to this, it is crucial to create a neural network design that accurately captures the features of transit journey time by using an effective prediction technique.

\subsection{Research Design}

For members of the supply chain and the efficiency of their logistics, it is extremely valuable to predict the journey times for freight transports. A material planner, for instance, can adjust material supplies, anticipate delayed deliveries, and change the capacity of employees and equipment. Improved transport planning and more precise capacity planning at their facilities are also advantages for logistic service providers. As a result, industries and logistic service providers can increase productivity, optimize workflow, and improve planning precision. In order to do this, a constant monitoring of freight movements is necessary, for example, utilizing mobile sensors attached to carried items (Fu et al., 2020).

Short-term forecasting rather than long-term forecasting is currently the focus of most published work. The majority of the time, research is conducted utilizing historical, statistical, Kalman filter, or machine learning (ML) methods to determine bus arrival times or highway traffic times. Most studies assert that the only models that can effectively handle the dynamic conditions that arise during transports are ML ones. The literature can attest to the truth of this assertion because ML techniques frequently outperform statistical and historical approaches. For instance, ML is better able to analyze complicated and noisy data and deal with predictor connections that are complex and non-linear (Abdollahi et al., 2020).

Even so, only a limited number of recent publications discuss the use of ML techniques like Convolutional Neural Network (CNN) or long short-term memory (LSTM) to forecast transit times in freight shipments (ANN). Only a little amount of study has been done on long-term forecasting of multimodal freight movements, and it primarily relies historical methods. As a result, the focus of this work is on the ML techniques in order to estimate the travel time in supply chain.

Machine learning is the study of statistically modeling a problem to carry out an action without having precise rules and instructions for doing so. When compared to conventional programming, it uses a series of intuitive techniques that reverses the order of tasks. In conventional programming, the desired result is calculated using a pre-established set of rules that are applied to the existing data (Chen et al., 2020).

In machine learning programming, pre-known data and the desired result are represented in a way that leads to the discovery of a previously unidentified set of rules. This comparatively new method of solving issues is very helpful in business in general but particularly in supply chain management. The task of modeling the supply chain using conventional approaches may be highly challenging or even impossible due to the sheer number of hidden and variable components that interact and correlate with the desired objectives in a complex ecosystem like the supply chain (Wu et al., 2020).

\subsection{Structure of the Paper}

The structure of this paper is as follows. Section~1 explains the background, aim, objectives and problem statement of the paper. In Section~2, the academic research available in the field of travel time prediction in supply chain management are explored and reviewed, to understand which machine learning techniques are currently in use for travel time prediction. A detailed literature review of related work is also done, to show the available research and what has been done in the field of travel time prediction. In Section~3, the methodology of this research will be discussed. Section~4 describes the results of all the applied machine learning methods for travel time predictions and contribution of this research in the field of travel time prediction. Section~5 explains the discussion, limitations and future work of this research. Section~6 details the summary, implications and recommendations from the research results.

\section{Literature Review}

\subsection{Background}

Today's freight transportation networks must connect increasingly distant production and sales regions, driven by global supply chains with a growing worldwide reach, and such global competitiveness leads to increased demands for service, delivery times, and cost-efficiency. Simultaneously, constraints such as limited facility space and laws (such as environmental protection and customs) must be taken into account (Ni et al., 2020). These variables increase the complexity and dynamics of global freight transportation networks, putting them at risk. As a result of escalating cost pressure, corporations aim to minimize risk buffers at the same time, resulting in greater vulnerability in supply chains. Companies prefer slightly slower transports with a consistent arrival time over short but unreliable transport times, according to interviews, because this allows them to develop solid procedures along the supply chain (Hou \& Edara, 2018).

This necessitates greater transparency throughout the entire manufacturing and transportation process. Transport operations are particularly important in this context since they do not take place in a controlled setting, such as a factory, but rather on a shared infrastructure that is subject to environmental impacts. For freight transportation, an accurate trip time prediction is extremely valuable since it helps supply chain players to improve the quality of their logistics (Sharma et al., 2020). A material planner at the receiving facility can predict delayed deliveries in advance, allowing for better material stock forecasting. Furthermore, a plant's capacity, such as employees or machinery, can be adjusted over time to improve efficiency. The same is true for logistic service providers. Staffing, ramps, and forklifts, among other things, can be arranged accordingly at warehouses, ports, and other hubs. As a result, manufacturers and logistics service providers may improve their productivity, streamline their processes, and improve the accuracy of their planning (Al-Sahaf et al., 2019).

A supply chain, in general, entails all collaborative operations by all participating organizations in the transformation of raw materials into the end product. This comprises obtaining raw materials, manufacture, assembly, and delivery to the ultimate client. To do so, logistics for material handling, transportation, and storage are required. Supply chain management, on the other hand, is concerned with the planning and management of supply chains, whereas logistics is concerned with the operational level (Cavalcante et al., 2019). The duty of providing information between the various phases of the supply chain is also part of logistics. Transportation is often handled by several corporations or organizations. These circumstances make the supply chain much more complicated, especially in multimodal transports with several transshipment sites. The goal of this research is to accurately estimate trip time in multimodal transportation. For effective administration of transportation operations and logistics in supply chains, accurate trip time estimations are critical (Priore et al., 2019).

\subsection{Research Objectives in Related Work}

The researchers Ashwini et al. (2022) have analyzed a comparison study that uses error metrics like Mean Absolute Errors (MAE), Median Absolute Errors (MdAE), Root Mean Squared Errors (RMSE), Mean Absolute Percentage Errors (MAPE), as well as R squared (r2 score) to find an appropriate model for assessing the effectiveness of linear as well as non-linear ML model for predicting travel time. It was discovered that the Random Forest Regression method works well for planners of transit operations when looking to optimize the timetable as well as other transit operations. But compared to other models, Random Forest is a sluggish and complicated approach.

The authors Yuan et al. (2020) implemented a Deep Neural Network (DNN) to create a dynamic bus journey time prediction system. The findings demonstrated that the algorithm's performance was improved by 4.82\% when compared towards the deep neural network used on the original feature space and was more effective than the conventional machine-learning model. However, a deep image convolutional network using reference image processing is required to implement this strategy.

The authors Servos et al. (2019), utilizing the machine learning techniques Extra Trees, AdaBoost, and SVR, have forecasted trip times for multimodal transportation. SVR has outperformed the other two used machine learning algorithms in terms of calculating journey time. Over a travel period of up to 30 days, the prediction error might be as much as 17 hours. However, the model will need to be retrained using newly completed rides in such an iterative process.

The researchers Viellechner and Spinler (2020) provided a container shipping sector between Europe and Asia a data analytics-based solution. The 315 explanatory variables, 10 regression models, including seven classification models that made up the prediction model. The neural network as well as support vector machine models, produced the greatest results using machine learning methods, with a prediction accuracy of over 77 per cent compared to a baseline model's prediction accuracy of 59 per cent. Even in the future additional real-time data should be included in a prediction model that focuses on daily short-term changes.

The authors developed an algorithm for XGBoost-based trip time prediction between bus stations. For training as well as verification, the 28-day bus operating data of a specific bus route in Guangzhou were utilized, and they were contrasted further with estimation techniques based upon K-Nearest Neighbor (KNN), BP neural networks, including Light Gradient Boosting Machines (LightGBM). The XGBoost prediction model was determined to have the least MAPE of 11.96\% when compared to other models, which would be 9.30\% lesser than other models generally. However, there is a need to increase the prediction's accuracy.

The authors have solely examined the methods in which projected time arrival could be predicted in order to establish a research need from the aforementioned studies. The majority of publications have utilized machine learning regression analysis models to determine the ETA; they have employed linear models, which have less capacity to predict the right ETA when given an intricate dataset. As a result, the suggested work may achieve a higher level of performance for prediction and is able to derive a decent forecast from a less quantity of data.

\subsection{Travel Time Prediction}

Continuous monitoring of freight transports, such as employing mobile sensors attached to move products, is essential for journey time calculation. Simultaneously, achieving the requisite transparency is one of the most difficult logistical jobs. Predicting travel time is especially challenging since several elements influence it, including weather, traffic, vehicle, route, and transportation relationship (Diez-Olivan et al., 2019). Today's freight transportation networks must connect increasingly distant production and sales regions, driven by global supply chains with a growing worldwide reach, and such global competitiveness leads to increased demands for service, delivery times, and cost-efficiency. Simultaneously, constraints such as limited facility space and laws (such as environmental protection and customs) must be taken into account. These variables increase the complexity and dynamics of global freight transportation networks, putting them at risk (Diez-Olivan et al., 2019).

In the framework of intelligent transportation systems, travel time information plays an essential role in transportation and logistics, and it has been employed in a variety of disciplines and for a variety of reasons. Accurate travel time prediction is critical when developing advanced traffic information systems (ATISs), which assist travelers in planning their routes ahead of time and guiding them along the way, as well as for traffic management issues such as traffic control system management and logistics planning and operations. Different approaches can be discovered in the literature from a methodological standpoint (Dogru \& Keskin, 2020). In a summary, short-term predictions based on data characterizing the traffic status are created using artificial intelligence or quantitative statistics approaches. We'd like to bring up data mining and pattern matching as examples of the first. In this paper, a machine learning strategy for predicting short-term trip times for a public mass transit firm is discussed. Linear regression models, support vector machines, and time series analysis are statistical methods that are appropriate for estimating journey durations over a limited time frame (Zantalis et al., 2019).

\subsection{Machine Learning in Travel Time Prediction}

Machine learning is the study of statistically modelling a problem in order to complete a task without having to specify particular rules and instructions. It's a set of approaches that, in contrast to traditional programming, rearranges the sequence of jobs. Traditional programming involves applying a well-defined set of rules to existing data in order to determine the desired result (Syafrudin et al., 2018). Pre-existing data and desired outcomes are represented in machine learning programming in order to develop a previously unknown set of rules. This relatively new technique to problem-solving can be tremendously beneficial in business, particularly in supply chain management. In a complex ecosystem like the supply chain, the amount of hidden and variable components that interact and correlate with the desired outcome can make standard modelling extremely difficult, if not impossible (Rolnick et al., 2019).

The majority of published literature currently focuses on passenger transportation, such as bus arrival times or highway travel times, rather than freight transportation (Tran et al., 2020; Balster et al., 2020). Only rides of up to two hours are considered in this case, which does not apply to freight transports. The authors employ average-based techniques, Kalman filters, or machine learning (ML) algorithms to estimate journey time. Furthermore, the route and stops are known ahead of time in their research (Konovalenko \& Ludwig, 2019). Only machine learning algorithms are capable of handling the complex and dynamic behavior that occurs during transportation. ML is stronger at dealing with complex and non-linear connections between predictors, as well as processing large amounts of data. This remark is supported by the research, which shows that ML techniques outperform average-based approaches in most instances (Wang \& Ross, 2018).

Despite this, machine learning has only been used in a small number of recent publications dealing with freight transportation journey time prediction. Only a small amount of research has been done on forecasting multimodal freight movements based on real-time tracking data, which mostly uses average-based algorithms. As a result, the focus of this research is on the evaluation of machine learning (ML) for long-term forecasting of multimodal freight transport transit times (Syam \& Sharma, 2018). Sensors affixed to the carried products create data that is used to track them. Due to limited battery life, sensor data is broadcast at a low frequency for 30 minutes at a time. Aside from the origin and destination, no further details about the route or trans-loading points are known ahead of time. Finally, we can demonstrate that, given the limits, machine learning systems are capable of accurately predicting travel time (Mercier \& Uysal, 2018).

Support vector regression (SVR) delivers the best prediction accuracy in this use-case, with a mean absolute error. It also demonstrates that the model outperforms average-based techniques. In order to automate routines, methods, and procedures in numerous fields of knowledge, new technologies, concepts, and processes in the field of Information Systems, as well as the specific branch of Computer Science, have been transformed into intense systems (Baryannis et al., 2019). The widespread adoption of these technologies has resulted in massive amounts of data. The knowledge of the organization, processes, and environments is embodied in these huge amounts of data. Data analysis procedures have given us access to this information. Database technology, on the other hand, is not sufficient to facilitate data analysis on its own. Complementary technologies and tools, primarily created in the domain of Data Mining and Machine Learning, are necessary for the procedure (Liu et al., 2021).

While there are other methods for estimating journey times, the majority of them are based on shallow learning architectures. In contrast to deep learning systems, deep learning architectures do not have the ability to learn features. A new journey time prediction technique can be done based on the Deep Belief Networks concept for novel research (DBN). A deep learning method has the advantage of being able to handle massive amounts of traffic data (Mohanta et al., 2020). In a general method, a stack of Restricted Boltzmann Machines (RBM) is used to automatically learn generic traffic features in an unsupervised fashion, and then a sigmoid regression is used to predict travel time in a supervised fashion. Unlike most travel time prediction methods which require the traffic data for each road link to be trained separately, for the method that can collectively train the traffic data on the entire road network all at once (Mart\'inez et al., 2020).

\subsection{The Travel Time Prediction Problem}

Travel demand and travel time variability in suburban and urban locations are significantly influenced by relative short-phased cyclic patterns, such as regular daily and weekly variations brought on by peak hours. The fluctuation is frequently greatly influenced by the short-phased patterns, particularly in urban and crowded locations. Weather, events, and more sporadic temporal variables like holidays are among the many other elements that are known to also have an impact on travel demand and journey time (Abdollahi et al., 2020).

Additionally, one of the biggest issues in contemporary society that hasn't been resolved is traffic congestion. Traffic congestion frequently happens because of changes in demand and poor forecasting, despite the fact that transportation infrastructure is built with consideration for both current and future traffic demand. It is difficult for transportation infrastructure, in particular motorways, to adapt to variations in traffic demand. It takes a lot of money and time to build a new highway or extend an existing one, and political and environmental concerns sometimes make it impractical. Consequently, it's crucial to make effective use of the current infrastructure (Fu et al., 2020).

It is helpful to know if an estimate is unclear due to large variances in historical data for comparable journeys since it suggests boosting the routing process's margins. There are databases that include travel times at the level of a road segment, which is ordinarily the section of a road in between intersections or traffic signals. Depending on the time of day, the day of the week, the legal speed limits, and previous data for each road segment, the trip times may also be differentiated (He et al., 2020). These maps provide a solid foundation for estimating journey times, but they are unable to reflect higher-level patterns, such as those that apply to individual trips (includes additional road segments).

There are some difficulties in predicting travel times. First of all, the complexity of traffic dynamics places significant demands on a model's ability to depict it. By taking into account vast amounts of historical data and applying machine learning-inspired techniques, the basic strategy is to let previous journeys speak as much as possible for themselves. Data sparsity may provide issues as it does with all data-driven methodologies. The inference requires a sufficient number of identical visits in order to get reasonable results (Fu et al., 2020). Henceforth, the sparsity problem is related to the concept of similarity since different approaches to similarity will be impacted by sparsity problems in different ways.

For instance, travels using origin-destination approaches must have comparable origins and destinations, whereas trips using path-dependent approaches merely need to use the same sub-paths. The sparsity problem can be summed up as a symptom of the curse of dimensionality, which refers to difficulties encountered while analyzing data with several dimensions. Dealing with complexity and scalability is the second difficulty. The size of the data sets being analyzed increases the computational complexity of many of the advanced methods in the literature significantly (Wu et al., 2020). As a result, greater care must be used while making estimates in order to simplify and speed up the algorithms.

\subsection{Estimated Time of Arrival (ETA) in Supply Chain Management}

Actors in the supply chain employ transport management systems and track-and-trace systems to improve their supply chain visibility. However, these systems currently only provide information on how a loading unit's transport is planned and where it is currently placed, not on how that unit's subsequent transport is likely to be achieved. This is especially crucial in the case of network outages that spread throughout the system (Nikolopoulos et al., 2021). A modest delay in the initial leg of the transportation chain might have detrimental cascade effects on all subsequent legs, finally leading to the missed of a critical planned connection. The estimated time of arrival (ETA) of transporters is an important metric that contributes to transparency. Intermodal transports, in which schedule-based and non-schedule-based transports are frequently integrated, are particularly interested in ETAs (Kilimci et al., 2019).

The term ``intermodal transport'' describes a transportation chain in which loading units such as intermodal containers are transported via at least two modes of transportation, including transshipment. The majority of the journey is usually covered by trains or ships that run on a fixed schedule. The loading unit's more flexible road transport is utilized only over short distances for pre- and post-carriage, such as transporting products to a rail or marine terminal or collecting items at the unloading point (Fei et al., 2019). Accurate and up-to-date ETAs for each transshipment point can be utilized in supply chains using intermodal freight transport networks (IFTN) to decide whether or not a connecting transport will be reached. If this were to be included in an information-sharing platform, proactive communication would be possible, allowing different players to consider and take necessary measures to potentially compensate for existing delays. By establishing ETAs, the resilience of IFTNs and supply chains can be improved. As a result, the various actors will be able to make the supply chain more efficient and cost-effective. A vast quantity of data with a wide product range and rapidity must be collected and analyzed in order to get reliable ETAs (Mao et al., 2018).

This is especially difficult for complicated IFTNs, as different parties are responsible for different parts of the intermodal transport chain. The work required for data gathering, cleaning and the connection is greatly increased by the various aims and IT systems. The biggest challenges in estimating ETAs in IFTN, however, originate from the differing characteristics of the various modes of transportation: the intermodal transport chain connects scheduled and unscheduled transports (Alizadeh et al., 2018). This results in peaks in the distribution of probable arrival times, preventing the establishment of an ETA prediction for the entire transportation chain with a given confidence interval. Furthermore, some modes of transportation, such as ships and railroads, have limited loading capacity, as well as transshipment capacity at transshipment sites, which adds to the difficulty of predicting ETAs (Abduljabbar et al., 2019).

These discussions were used to determine the requirements for ETA prediction, as well as the available data, potential transportation disruptions, and anticipated problems in implementing integrated ETA prediction for entire multimodal transport chains. Process data from railway operators, railway transport companies, and inland terminal operators, network data from railway infrastructure firms, and supplementary weather data from weather services are all included in the data. When significant volumes of data are available, one option is to create a descriptive microsimulation model that depicts the underlying system's structure in great detail (Milojevic-Dupont \& Creutzig, 2021). The magnitude of the dynamics and complexities within IFTNs, on the other hand, makes such a model impractical. Nonetheless, machine learning (ML), a new subject that has lately emerged, offers new opportunities. Because intermodal transportation chains sometimes have buffer hours between the rigorous timetables of some modes of transportation and often include extra services, such as storage, a solely data-driven method utilizing ML would not be appropriate for ETA forecasts in IFTNs (Kannangara et al., 2018).

\subsection{Machine Learning Approach Estimation of Travel Time}

The researcher (Cheng et al., 2019) explained that when machine learning is applied to the entire intermodal transportation chain, it results in a prediction model that ignores the logistics structure of the chain and so blends transport times, buffer periods, and storage times. Because it would be impossible to discriminate between scheduled and unforeseen storage times, the ETA estimate provided in this method would be meaningless. In addition, the authors (Fu et al., 2020) discussed the same origin-destination pair, a container can be sent via several routes. If users eliminate this characteristic, the projections will be drastically different. To construct an acceptable ETA prediction model for an intermodal transport chain using machine learning, knowledge of the actual transport processes is required. This understanding allows for the identification and incorporation of logistical structures into the overall strategy.

Other authors (Abdollahi et al., 2020) stated that there is no research that uses a combination of machine learning approaches and logistic structure mapping to forecast ETA in IFTNs. In recent years, freight transportation research on ETA predictions has primarily concentrated on single types of transportation, particularly truck transport. As a result, a large portion of the data produced by today's IFTNs is useless. In terms of openness and operational efficiency, being able to exploit these data through the new possibilities given by ML has a ton of potential. The authors (Miao et al., 2020) explained that the goal of the computational study is to propose an approach to ETA prediction that takes into account available data from various actors and covers the full intermodal transportation chain. The overall ETA prediction was separated into subproblems covering the individual legs of the intermodal transport chain, and an appropriate ML method was selected for each leg to decrease the complexity and make specific and accurate predictions for each person.

Some researchers (Ma et al., 2019) described the methods to predict the time in four steps, each sub model was developed: (1) system structure, (2) feature engineering and feature selection, (3) model selection and model tuning, and (4) system validation. Because the data allows for the identification of individual containers as well as the assignment of trucks and wagons on a train, all predictions may be transferred to successive legs of the transport chain and used as inputs for subsequent forecasts. As a result, all of the individual estimates may be merged into a single overall ETA forecast that encompasses the whole intermodal transportation chain from origin to destination.

\subsection{Prediction of Time in Other Sources}

According to researchers (He et al., 2019), for a variety of reasons, predicting journey time is challenging. Traffic conditions or the speed changes of the cars along the section must be estimated in order to anticipate travel time for a piece of road. Because traffic conditions can change greatly in both space and time, it is difficult to precisely anticipate them. However, a significant amount of study has been done in this field's related fields. Because various means of transportation operate on rigid schedules, the question of whether a container can make its scheduled connection on time arises. The authors (Philip et al., 2018) proposed that the flexible container delivery by truck in the hinterland terminal with onward transportation by scheduled train, the formation of a scheduled train in the marshalling yard for onward transfer to the port of Hamburg, and the marshalling of wagons from the port's entry station to the sea terminal for loading onto the scheduled ship are the most important of these transitions. The first leg of the total shipment is frequently a truck because the shippers of the containers at the starting node are enterprises that rarely have a direct relationship to trains.

Researchers (Ran et al., 2019) explained that the most important of these transitions is the flexible container delivery by truck in the hinterland terminal with onward train transportation, the formation of a scheduled train in the marshalling yard for onward transfer to the port of Hamburg, and the marshalling of wagons from the port's entry station to the sea terminal for loading onto the scheduled ship. Because the shippers of the containers at the starting node are typically businesses with no direct relationship to railroads, the first leg of the whole shipment is frequently a truck.

The author (Crist\'obal et al., 2019) studied that for all kinds of transportation when it comes to ETA estimates. Many ETA prediction techniques for road traffic, such as navigation systems, have been applied in practice. These solutions, on the other hand, are frequently inadequate to meet operational needs or to account for dynamic occurrences like the weather. Furthermore, most existing solutions do not contain ETA-based real-time operations management metrics. The scientific literature and practical solutions for rail transportation are primarily focused on passenger transport, while rail freight transport is rarely considered.

According to research by authors (Wang et al., 2018) which used data from loop detectors and a global positioning system (GPS) to calculate highway speed, occupancy, and volume, as well as prior trip times, to anticipate travel times over a four-mile stretch of roadway. As a learning algorithm, they employ artificial neural networks (ANNs). Similarly, the authors use support vector machines and loop detectors to estimate trip time on highway stretches up to 350 km across.

Some researchers (Kumar et al., 2019) identified that random forests (RF) as the best approach for predicting travel times on urban street segments using GPS data from 300 probe vehicles. In a comparable application, the authors compare the performance of a gradient boosting method and RF for estimating trip time on highway segments and find that the gradient boosting method outperforms the RF method. Gradient boosting also yields an excellent forecast, according to the authors. The authors also use GPS data to analyze several aspects. They discovered that factoring in the speed difference between two broadcasts improves prediction accuracy by 5\%.

The authors (Zhao et al., 2018) used 12 months of GPS data and the prior 12 travel times as features to analyze deep neural networks. Their method yields a mean absolute error (MAE) of 3.25 minutes over a 131-kilometre length. As with freight transportation, the preceding methodologies do not take into account any end-to-end transit relationships. Furthermore, precise route knowledge and a high frequency of GPS measurements are necessary, otherwise, side-based techniques such as loop detectors must be used. ANNs are used by practically all writers in the prediction of bus travel times. The authors (Yang et al., 2018) have also used ANNs in tramways. In all circumstances, the route is known ahead of time and is divided into sections based on the bus stops. The majority of ways merely differ when it comes to the analyzed use-case and the features that were utilized. Time of departure, public holidays, dwell time at bus stations, current travel time, distance to destination, and average speed are typically used.

Authors (As \& Mine, 2018) proposed ANNs over k-nearest neighbors (knn), but show that support vector regression (SVR) performs better. For prediction, the authors utilize a multilinear regression model. The authors include the total number of stops and dwell time to their model to include all previous stops. In contrast to earlier techniques, the bus ride is linked to an origin and destination via an end-to-end transportation relationship that includes pauses, much like freight logistics. The authors, on the other hand, have not taken into account that numerous stops in supply chains are typically unknown to supply chain participants and that transports using the same transport connection can differ. Furthermore, their methods necessitate a large number of rides and rely on frequent GPS data.

\subsection{Research Gap}

However the authors (Kumar et al., 2019) have discussed about the possibilities, few limitations can be reduced with a use of a practice-oriented application of machine learning in order to increase transport network dependability. Also as per the researchers (Zhao et al., 2018) the findings demonstrate the necessity of considering logistic nodes when predicting ETAs in intermodal transportation, as well as the importance of having information about the logistics process, even if it is not modelled in every detail. Henceforth, the findings demonstrate the importance of data availability and quality. Organizations will need to concentrate on data availability and quality first if they wish to make greater use of their existing capacities in the future utilizing ML algorithms. Also the authors (Fu et al., 2020) stated that the operators in multimodal transportation chains can be used to create models to assess the immediate impact of delays on downstream activities which can be improved in the future. Thus, the ETA data is largely used as a tool for providing early support for operational decision-making issues, such as people, vehicles, tools, and infrastructure disposition. Henceforth, predicting travel time using machine learning helps to alleviates traffic congestion and improves operation efficiency and utilizing ML can give better accurate results.

\subsection{Conclusion}

The Expected Time of Arrival (ETA) helps to measure the travel time between two points of origin and destination. It's a crucial location-based service for interactive maps and navigation. ETA has a lot of uses on the ride-hailing platform, since travel time is one of the most important factors for drivers and riders when making a contract. As a result, it is important to predict travel time accurately before venturing on a trip. A reliable ETA would improve the transportation system's effectiveness, lowering consumers' travel costs, reducing electricity demand, and lowering emissions from motor vehicles. As a result, ETA has become a key factor in decision-making at various stages of the online ride-hailing process, such as route discovery and vehicle dispatch (Jian Li, 2020).

A novel machine learning algorithm can be used to correctly estimate the travel time along a specific route at a specific departure time (Wang et al., 2018).

\section{Methodology}

\subsection{Methodology}

The following can be used to describe the research's methodology. The dataset for this investigation is first examined using web resources (variables like shipment detail, Shipment ID, source locations, destination locations, latitudes, longitudes, temperature, humidity, rains, expected time of arrival, distances, etc.). Using exploratory data analysis, it is possible to further evaluate numerous aspects that might be related to variations in trip times in supply chain management. The data must then be pre-processed. Pre-processing is applied to reduce the amount of noise in the data and eliminate null and missing values.

Principal component analysis or the random forest algorithm are then used to identify significant elements in the dataset that relate to ETA variations. Other techniques employed by the authors include lasso regression (Viellechner \& Spinler, 2020) and clustering (Servos et al., 2019). The majority of machine learning algorithms have a tendency to be linear models, which have significant drawbacks when dealing with complicated datasets as well as make achieving lower loss rates challenging (Viellechner \& Spinler, 2020).

Artificial neural networks function like a perfect nonlinear model, however, they are ineffective for complicated datasets with plenty of features but small sample sizes. So, in order to conduct regression tasks, we need to look at techniques like recurrent neural networks or long short-term memory and then apply 1D-CNN, which makes use of the correlation between features, to interpret the data. The chosen features are then used to a 1D-CNN to create feature maps, which are representations of the data that contain significant information. And the long short-term memory (LSTM) method is then used to process those feature maps to produce the final result, which is the shipment's projected arrival time.

\subsection{Outline of This Section}

\begin{itemize}
  \item Data collection for estimated time arrival in supply chain management
  \item Data pre-processing
  \item Research framework
  \item Dimensionality reduction or feature selection
  \item Principal component analysis
  \item Deep learning modelling for estimated time arrival prediction
  \item One dimensional convolutional neural network (1D-CNN)
  \begin{itemize}
    \item Convolutional layer
    \item Pooling layer
    \item Fully connected layer
    \item Activation function
  \end{itemize}
\end{itemize}

\subsection{Data Collection for Estimated Time Arrival in Supply Chain Management}

Without an efficient and reliable method for gathering the data, it will be impossible to conduct data-based analytics or make data-based decisions in the future. Although there are many ways to collect data, including Auto-ID technologies, sensor technologies, digital gadgets, and social systems based on the Internet, there are still some difficulties in the supply chain and manufacturing sectors (Zhong et al., 2016). Transferring goods from one part of the world to other is an important process when it comes to supply chain management. Equipment used in this transferring process may include trucks, trains, flight and shipment via water. For our study, the main foundation is to collect accurate data of shipment delivery, which contains information about geo location variables, time variables, weather variables and shipment data.

As discussed above there are several ways in which data can be collected; for the shipping sector one of the ways the data is collected is through sensors and GPS technology (Servos et al., 2019). Authors have carried out the process using 43 pallets that have been distributed among seven container shipments. The container's pallets are individually fitted with a sensor. From a tag attached to the palette, the sensor is linked to transport-related data, such as the serial number. When the packet is scanned, the palette is activated to start the process. The sensor uses Bluetooth Low Energy to send quality-related data, such as temperature and humidity, along with its ID. This data was used by the authors to predict the estimated time prediction for shipment delivery.

For our study, we will be considering various online sources where shipment details are being recorded. Departure and arrival time of the shipment are noted for a particular time period of certain range of locations. These details are collected from Eesea (2022), which is a maritime intelligence website that lends data about shipment details as paid services. A considerable amount of data is then collected for year 2023. Then we have gathered further information using Automatic Identification System (AIS), which is a non-public data which stores information about ports and track vessels in the shipment. We have collected AIS data from datalastic.com (2024), which contains the shipment details, shipment ID, distance travelled etc.

\subsection{Data Pre-processing}

Data cleansing is the important part of developing a successful artificial intelligence model. Prior to attempting to analyze the case study data, we need to evaluate and resolving data quality issues in order to remove any defect in data (Kantardzic, 2011). While preprocessing the data, one needs to be careful to keep essential information from being lost while keeping the data clean enough to allow the finding of useful information patterns.

In our case study, we have collected 2 years of shipment delivery activity including delivery dates, distances, geo locations, and weather details. Although the data was extracted from actual delivery activities, some of the data can be incomplete or can contain duplicates which might cause problem when performing modeling on deep learning algorithms. So, we have removed features containing more null elements and filled features with less element using mean and median values. We have also removed duplicate records which was present in the dataset. Some information gathered from source may be outliers and distorted. Thus, we identified those outliers in the dataset using various statistical and data visualization techniques and removed those if necessary.

\subsection{Research Framework}

\begin{figure}[htbp]
  \centering
  \includegraphics[width=0.95\textwidth]{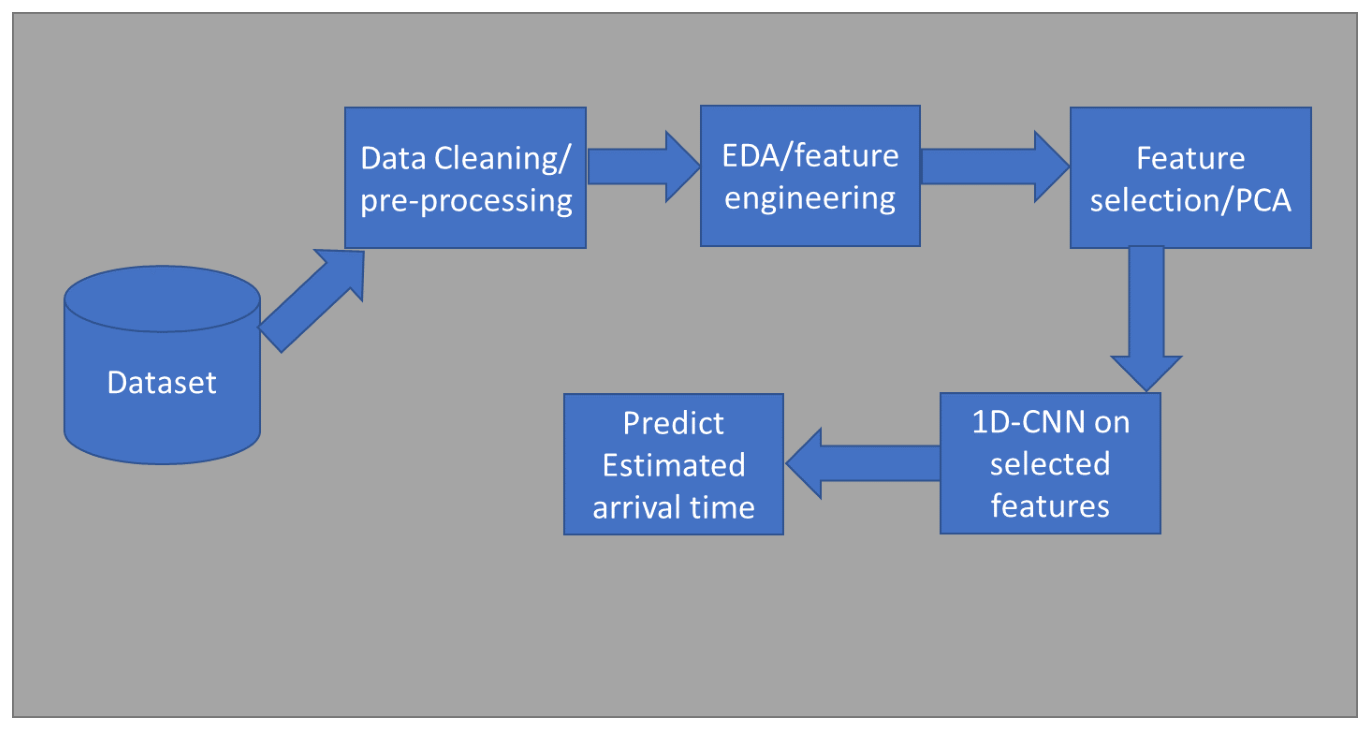}
  \caption{Model framework}
  \label{fig:framework}
\end{figure}

The framework of the work is given in Figure~\ref{fig:framework}.

\subsection{Dimensionality Reduction or Feature Selection}

In the realms of scientific study and industrial production, high-dimensional data is pervasive. Although it provides a wealth of information to users, its sparseness and repetition pose significant difficulties for data mining and pattern identification. Feature reduction can lessen redundant information and noise, simplify learning methods, and increase classification accuracy. This technique for minimizing the amount of input variables in training dataset is referred to as dimensionality reduction. After preprocessing, we further delete a few variables to speed up learning, decrease computation time, and make the model simpler without affecting performance.

\subsection{Principal Component Analysis}

We will be using principal component analysis as a dimensional reduction technique to remove unwanted features and retain important features with more than 95\% variance in the data. With a high number of dimensions or features per observation, principal component analysis (PCA) is a common method for analyzing huge datasets, improving data interpretation while retaining the most information, and enabling the visualization of multidimensional data. After performing PCA, the output data will be in the form components and the amount of components can be set before modeling and it can also be used as a hyper parameter for training the model (Karamizadeh et al., 2013).

We can adjust the number of components to observe which one predicts more accurate with lesser error rate. PCA is used to create predictive models and for exploratory data analysis. It is frequently used to acquire lower-dimensional data while retaining the most of the data's variation by projecting each data point onto the first few principal components. The path that maximizes the variance of the predicted value can also be used to define the first main component (Gewers et al., 2022). The direction that maximizes the variance of the projected data and is orthogonal to the first $i-1$ principal components is the $i$-th principal component. The dimension reduced data is then moved to 1D-CNN.

\subsection{Deep Learning Modelling for Estimated Time Arrival Prediction}

Deep learning describes a family of learning algorithms rather than a single method that can be used to learn complex prediction models, e.g., multi-layer neural networks with many hidden units. Deep learning has been successfully applied to several application problems in wide variety of sectors all around the world. There are various categories of algorithms when it comes to deep learning, such as Deep neural network (DNN), Convolutional neural network (CNN), Recurrent neural network (RNN), Deep auto-encoder (DAE), Restricted Boltzmann Machine (RBM), Generative adversarial network (GAN), and Deep reinforcement learning (DRL) (Hosseinnia Shavaki \& Ebrahimi Ghahnavieh, 2022). One of the few algorithms which have been appearing in recent years is LSTM, an advanced version of RNN which retains more memory of the sequence of features sent when training the model (H\"opken et al., 2021).

\subsection{One Dimensional Convolutional Neural Network (1D-CNN)}

\begin{figure}[htbp]
  \centering
  \includegraphics[width=0.55\textwidth]{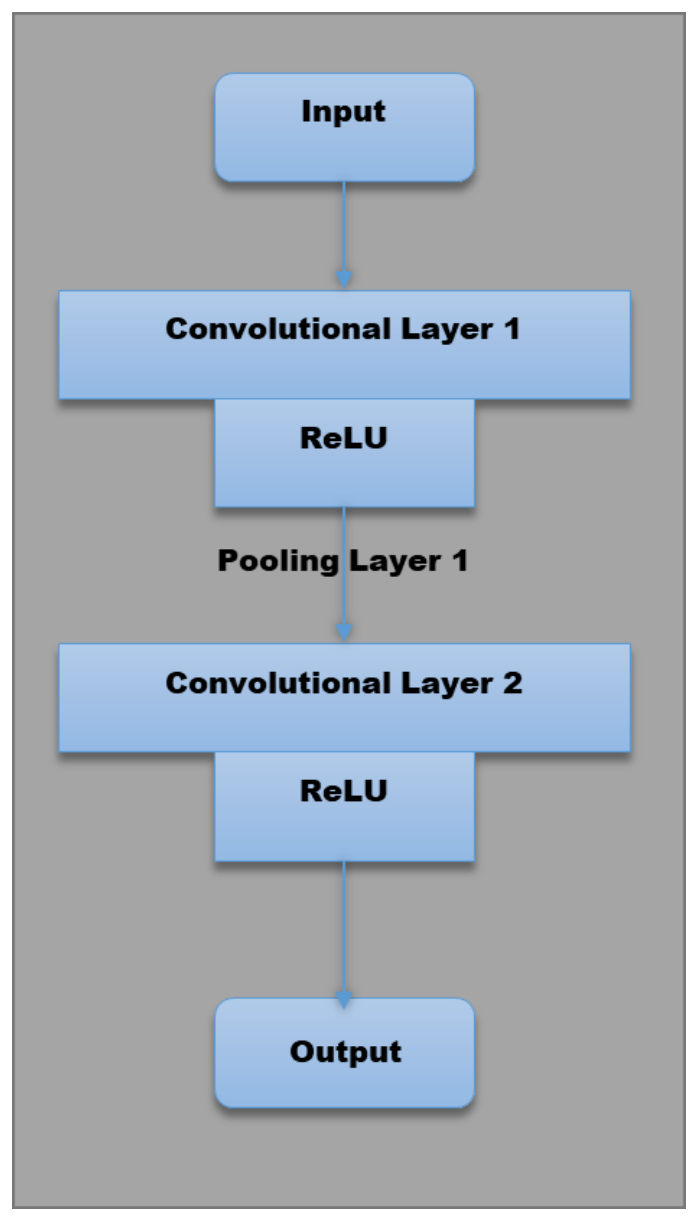}
  \caption{1D-CNN network structure}
  \label{fig:cnnstructure}
\end{figure}

The ``receptive field'' feature is added to the neural network to form the convolutional layer. LeNet, the initial model of the Convolutional Neural Network (CNN) model, was inspired by the study of the cat's visual system in neuroscience. The CNN model currently offers great accuracy and a quick training rate (Kiranyaz et al., 2021). The primary purpose is to draw features from the data. Its structure, which is depicted in Figure~\ref{fig:cnnstructure}, consists of a convolutional layer, an activation layer, a pooling layer, and a fully connected layer.

\subsubsection{Convolutional Layer}

The key to extracting features is the convolutional layer. Extraction of the corresponding data characteristics is carried out via the convolution kernel. The retrieved features become more abstract as there are more convolution kernels. Two convolutional layers are cascaded in the 1D-CNN network used in this article. Each convolutional layer's output is triggered by the nonlinear function ReLU and used as the input for the subsequent convolutional layer.

\subsubsection{Pooling Layer}

There are two types of pooling: maximal pooling and average pooling. The maximum pooling approach is used in this paper, and some unnecessary features are disregarded using the ReLU activation function. The final prediction result is obtained by the fully connected layer using a nonlinear function, while the ReLU activation function is a piecewise linear function (Wang et al., 2021).

\subsubsection{Fully Connected Layer}

The fully connected layer's job is to multiply the aggregated neurons into a one-dimensional vector form so that the data may be processed more quickly. In order for the convolutional layer and the pooling layer to learn the ideal parameter matrix, the weight update of CNN employs the back propagation technique of error to continuously alter the network connection weight to minimize the error.

\subsubsection{Activation Function}

\begin{figure}[htbp]
  \centering
  \includegraphics[width=0.45\textwidth]{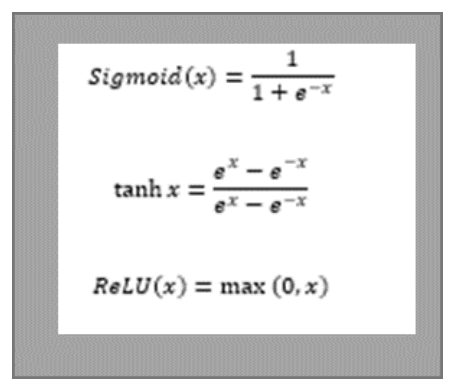}
  \caption{Activation functions}
  \label{fig:activations}
\end{figure}

The primary job of the activation function is to execute some nonlinear mapping on the features that the convolutional layer extracted in order to enhance CNN's capacity to handle certain nonlinear data. The ReLU, Sigmoid, and Tanh activation functions are the three most often utilized activation functions. The structure for solving nonlinear problems includes an important component called the activation function, which plays a significant role in maintaining features and eliminating redundant information (Wang et al., 2021). Figure~\ref{fig:activations} illustrates the functional relationship between Sigmoid, tanh, and ReLU.

Recently, 1D-CNNs gained attention, and they quickly attained the best performance levels in a number of applications, including the classification and early diagnosis of personalized biomedical data, the monitoring of structural health, the identification and detection of anomalies in power electronics, and the detection of electrical motor faults. Another significant benefit is the feasibility of a real-time and affordable hardware implementation due to the straightforward and compact setup of 1D CNNs that only carry out 1D convolution (Li et al., 2020b).

Structurally, 1D-CNN is almost the same as CNN, which also includes a series of convolutional layers and pooling layers, and finally outputs the results through a fully connected layer. In terms of usage, CNN is mainly employed for feature recognition of two-dimensional images, while 1D-CNN is widely adopted for feature recognition and extraction of time series. Although 1D-CNN has only one dimension, it also has the advantages of CNN's translation invariance for feature recognition. In 1D-CNN, since the convolution kernel is one-dimensional, a large convolution kernel will not bring too many parameters and calculations (Kiranyaz et al., 2019).

A larger convolution kernel can be used by the model to create a broader receptive field and more thoroughly extract the sequence's feature value. The input layer of the 1D-CNN network receives the distribution feature vector as an input. Features are extracted for each subgroup using the 1D-CNN network (Li et al., 2019). Convolutional layers are cascaded in the 1D-CNN network used in this study. The non-linear function ReLU serves as the input of the pooling layer and activates the output of each convolutional layer. A maximum pooling layer is implemented in between convolution processes to avoid over-fitting and enhance operational efficiency.

In our study, we will be passing the data that has been selected using dimensionality reduction process to 1D-CNN to get feature maps that contains important information out of the data. This then will be passed to the deep learning regression module to get the prediction of estimated time of arrival of shipments.

\section{Results}

\subsection{Exploratory Data Analysis (EDA)}

\begin{figure}[htbp]
  \centering
  \includegraphics[width=0.55\textwidth]{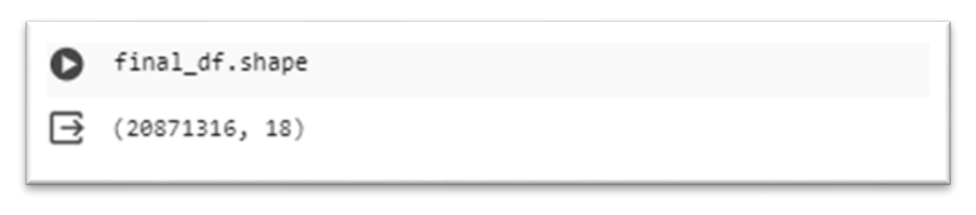}
  \caption{Shape of input data frame}
  \label{fig:shape}
\end{figure}

In this section, we study the data collected and find features that better explain the regression problem target. The data frame of the travel times and co-ordinates has over twenty million records with 18 features (Figure~\ref{fig:shape}).

\begin{figure}[htbp]
  \centering
  \includegraphics[width=0.9\textwidth]{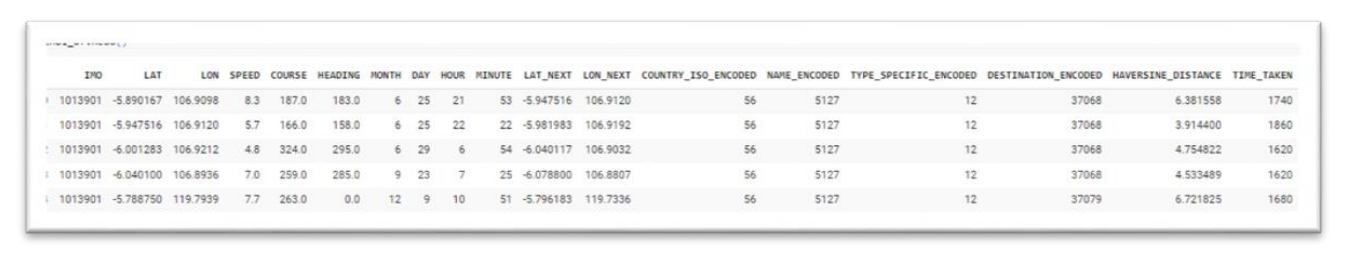}
  \caption{Sample records}
  \label{fig:sample}
\end{figure}

This is stored in a Pandas data frame. All the features have been converted to numerical ready for modelling. The features \texttt{COUNTRY\_ISO}, \texttt{NAME}, \texttt{TYPE\_SPECIFIC} and \texttt{DESTINATION} have been encoded using label encoding (Figure~\ref{fig:sample}).

\begin{figure}[htbp]
  \centering
  \includegraphics[width=0.5\textwidth]{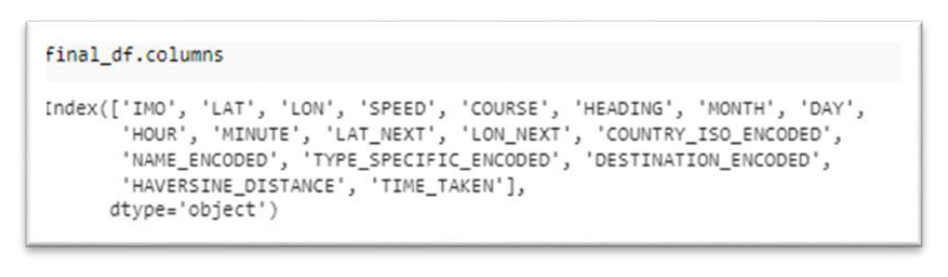}
  \caption{Input features}
  \label{fig:features}
\end{figure}

There are 18 features in the dataframe. All of them seem relevant to the target feature -- \texttt{TIME\_TAKEN} (Figure~\ref{fig:features}).

\begin{figure}[htbp]
  \centering
  \includegraphics[width=0.75\textwidth]{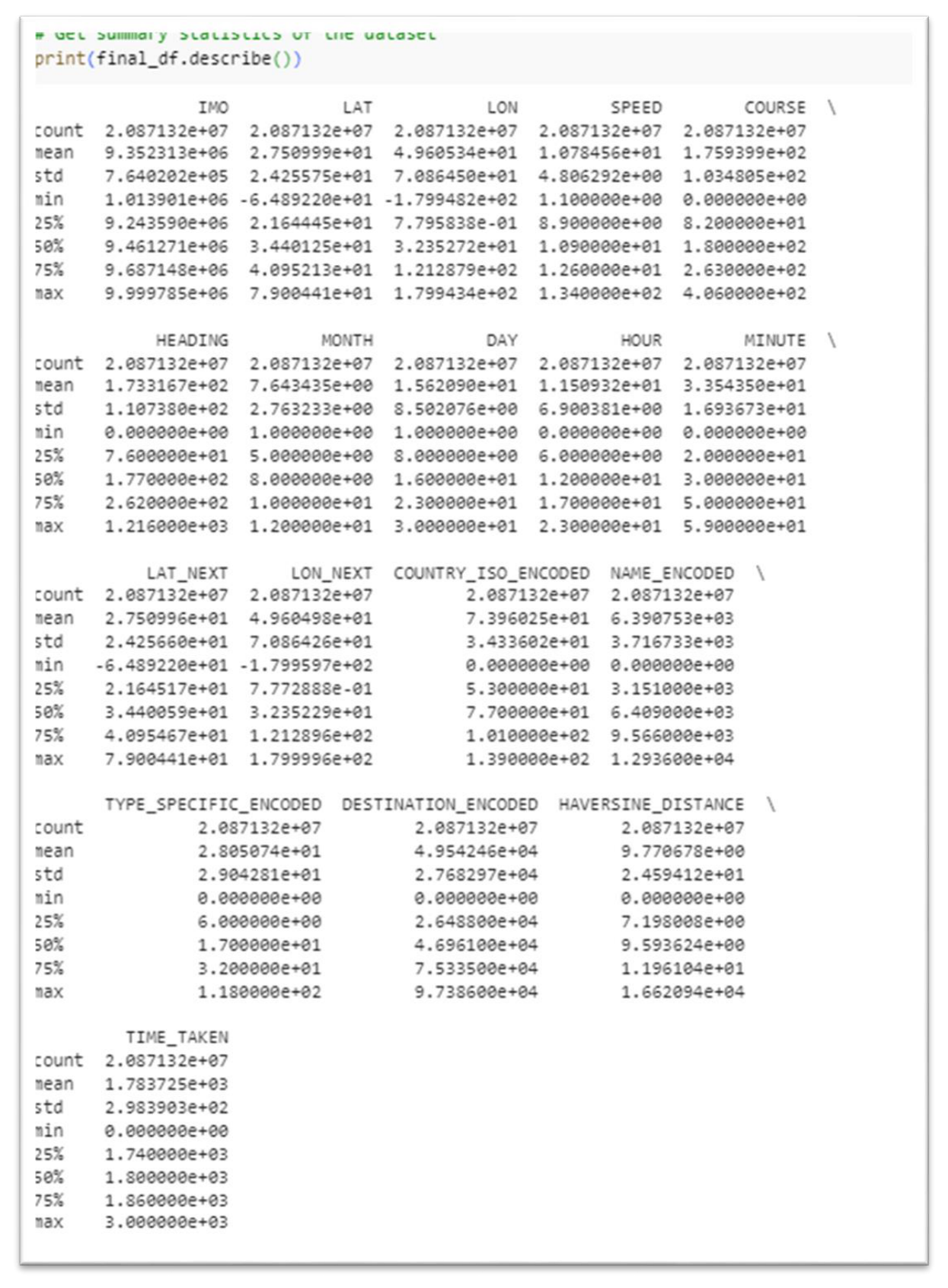}
  \caption{Descriptive statistics of input data}
  \label{fig:describe}
\end{figure}

Figure~\ref{fig:describe} shows the summary stats of the data. Each feature has its count, mean, standard deviation, minimum, first quartile, median, third quartile and maximum values. We will study this further in the data visualization part.

\begin{figure}[htbp]
  \centering
  \includegraphics[width=0.4\textwidth]{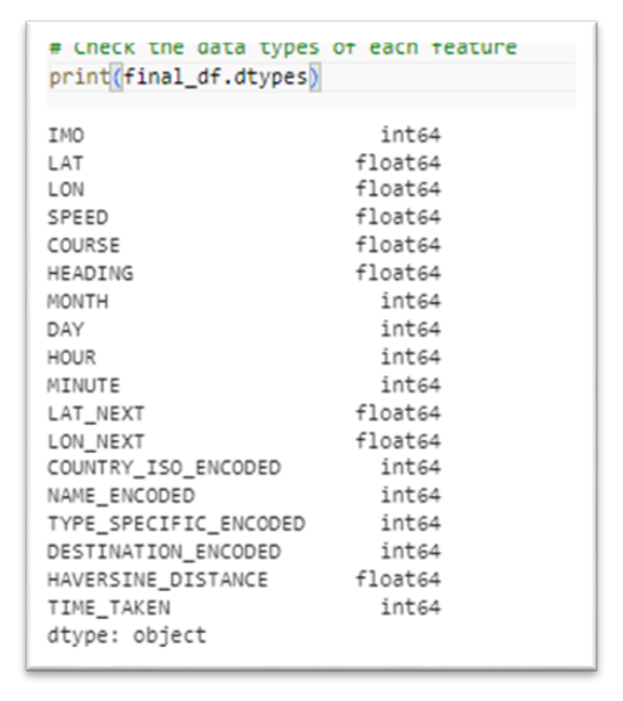}
  \caption{Input data datatypes}
  \label{fig:dtypes}
\end{figure}

\begin{figure}[htbp]
  \centering
  \includegraphics[width=0.35\textwidth]{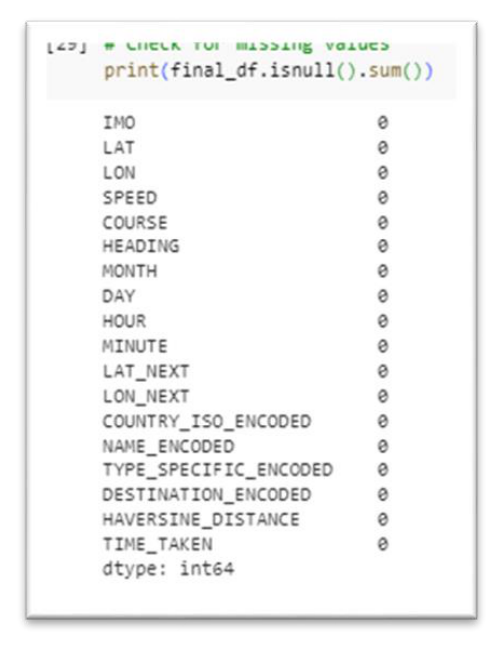}
  \caption{Input data missing values check}
  \label{fig:nulls}
\end{figure}

When we check the data types, all features are either integer or float and good for machine learning/deep learning modelling (Figure~\ref{fig:dtypes}). There are no NULL values present in the dataset (Figure~\ref{fig:nulls}).

\begin{figure}[htbp]
  \centering
  \includegraphics[width=0.95\textwidth]{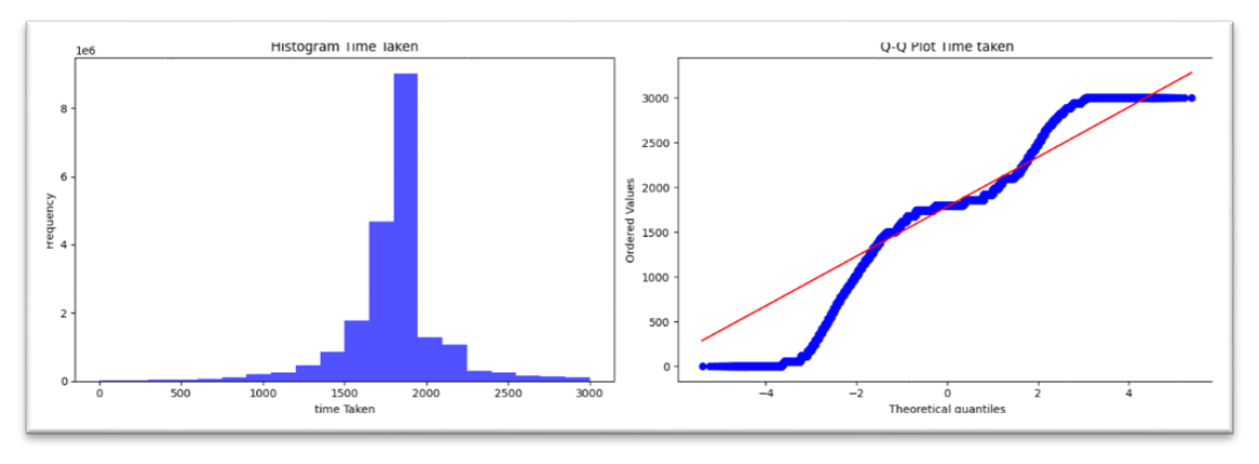}
  \caption{Target variable histogram and Q--Q plot}
  \label{fig:target}
\end{figure}

The target variable \texttt{TIME\_TAKEN} shows a distribution as in Figure~\ref{fig:target}. Most of the values are around the range 1500--2000 seconds. This is as per the data collected and can be used for predictive modelling. The Q--Q plot shows that the distribution is close to a normal distribution.

\begin{figure}[htbp]
  \centering
  \includegraphics[width=0.85\textwidth]{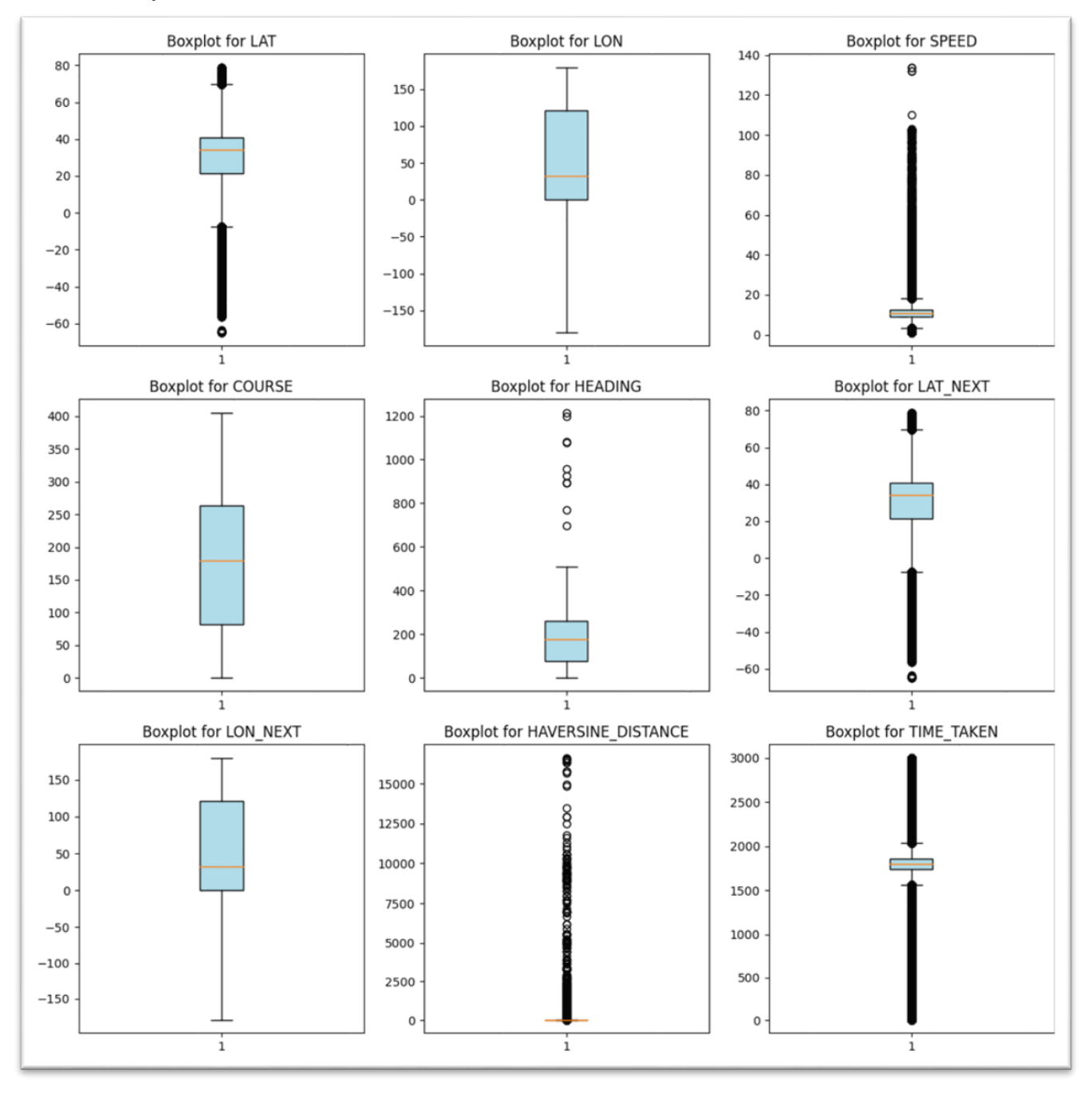}
  \caption{Box plot of prominent features}
  \label{fig:boxplot}
\end{figure}

Next we study the box-plot for 9 of the important features (Figure~\ref{fig:boxplot}). Below are the observations:
\begin{itemize}
  \item Feature \texttt{LAT} -- there are large number of outliers on the lower whisker. This means there are less number of occurrences of these latitudes in the data.
  \item Feature \texttt{LON}, \texttt{COURSE}, \texttt{LON\_NEXT} -- the data seems to be evenly distributed within the range of $-150$ and $150$.
  \item Feature \texttt{LAT\_NEXT}, \texttt{SPEED}, \texttt{HAVERSINE\_DISTANCE}, \texttt{TIME\_TAKEN} -- very randomly distributed. Lot of values are not repeating and the range is high for \texttt{HAVERSINE\_DISTANCE}.
  \item Feature \texttt{HEADING} -- narrow distribution range and lot of outliers on the upper whisker.
\end{itemize}
We will study the data further and do PCA to avoid the uneven distributions.

\begin{figure}[htbp]
  \centering
  \includegraphics[width=0.85\textwidth]{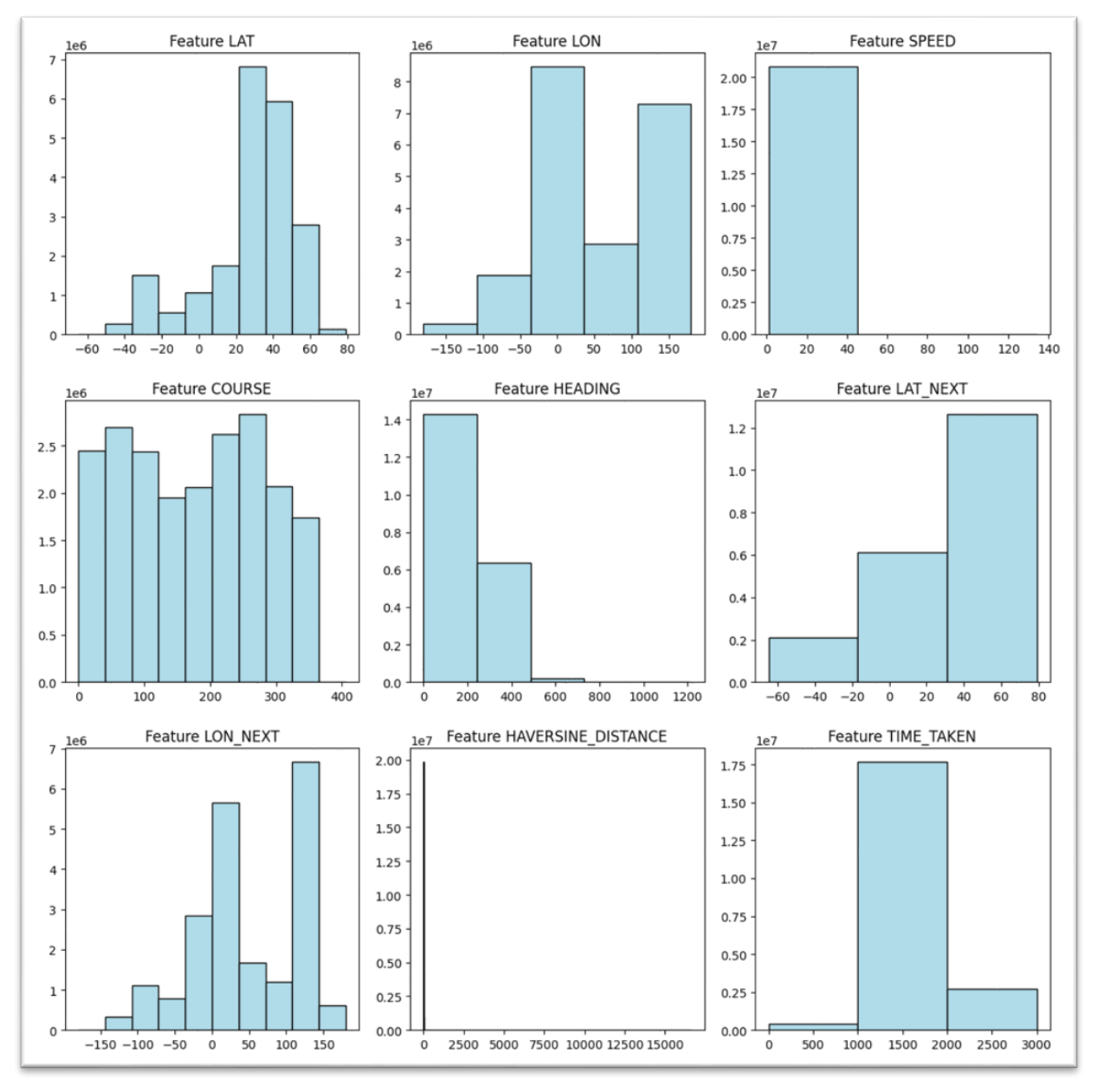}
  \caption{Histogram of prominent features}
  \label{fig:hist}
\end{figure}

The bar plots with the bins show how the range values where the frequencies are high (Figure~\ref{fig:hist}). Out of this, \texttt{TIME\_TAKEN} seems to have a normal distribution and \texttt{HAVERSINE\_DISTANCE} has a very narrow set of values.

\begin{figure}[htbp]
  \centering
  \includegraphics[width=0.8\textwidth]{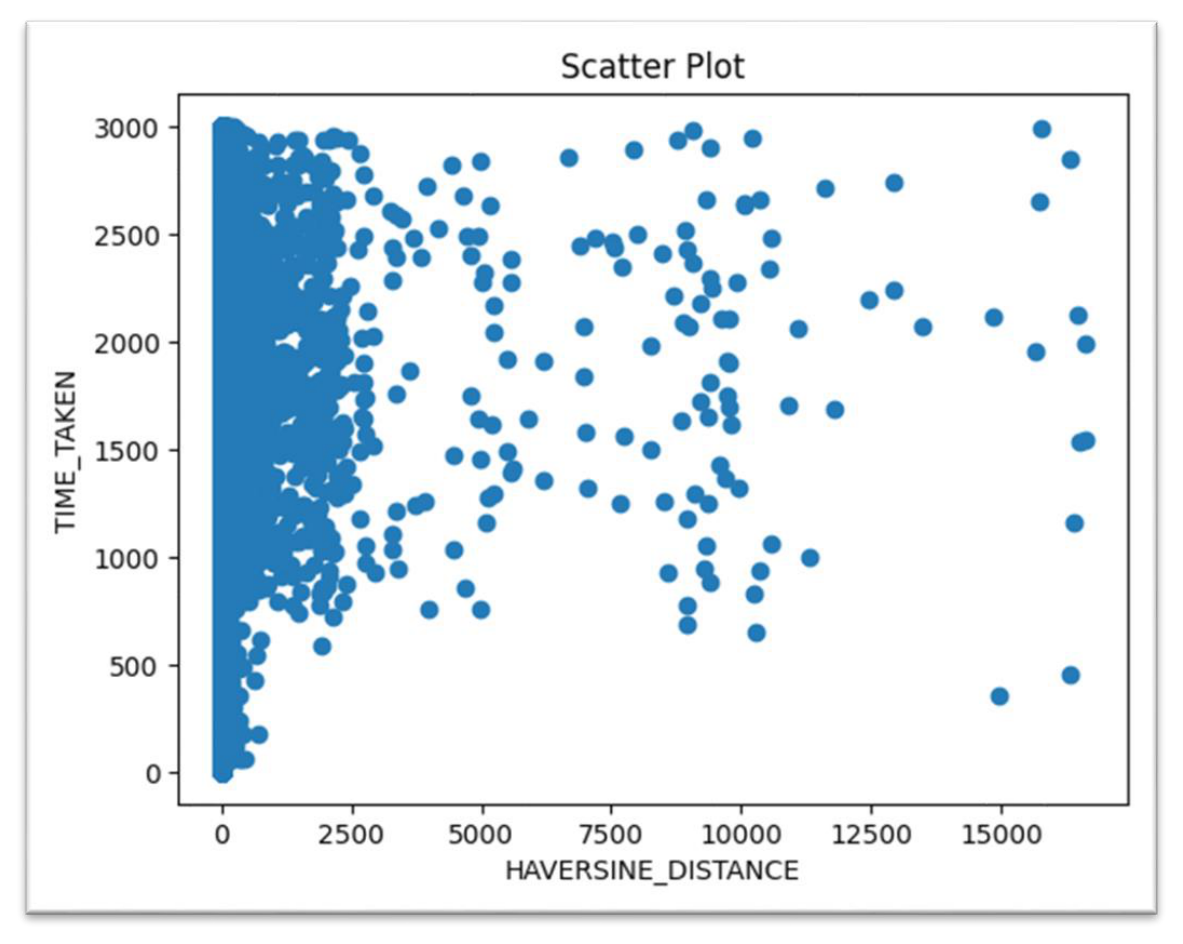}
  \caption{Scatter plot of \texttt{TIME\_TAKEN} vs \texttt{HAVERSINE\_DISTANCE}}
  \label{fig:scatter}
\end{figure}

The scatter plot of \texttt{HAVERSINE\_DISTANCE} vs \texttt{TIME\_TAKEN} (Figure~\ref{fig:scatter}) shows that the \texttt{TIME\_TAKEN} and \texttt{HAVERSINE\_DISTANCE} are highly correlated for all \texttt{TIME\_TAKEN} values when \texttt{HAVERSINE\_DISTANCE} is $\leq 2500$. This is meaningful as the distance covered varies along with speed of the vessel and conditions during the snapshot time and location.

\subsection{Correlation Study}

\begin{figure}[htbp]
  \centering
  \includegraphics[width=0.95\textwidth]{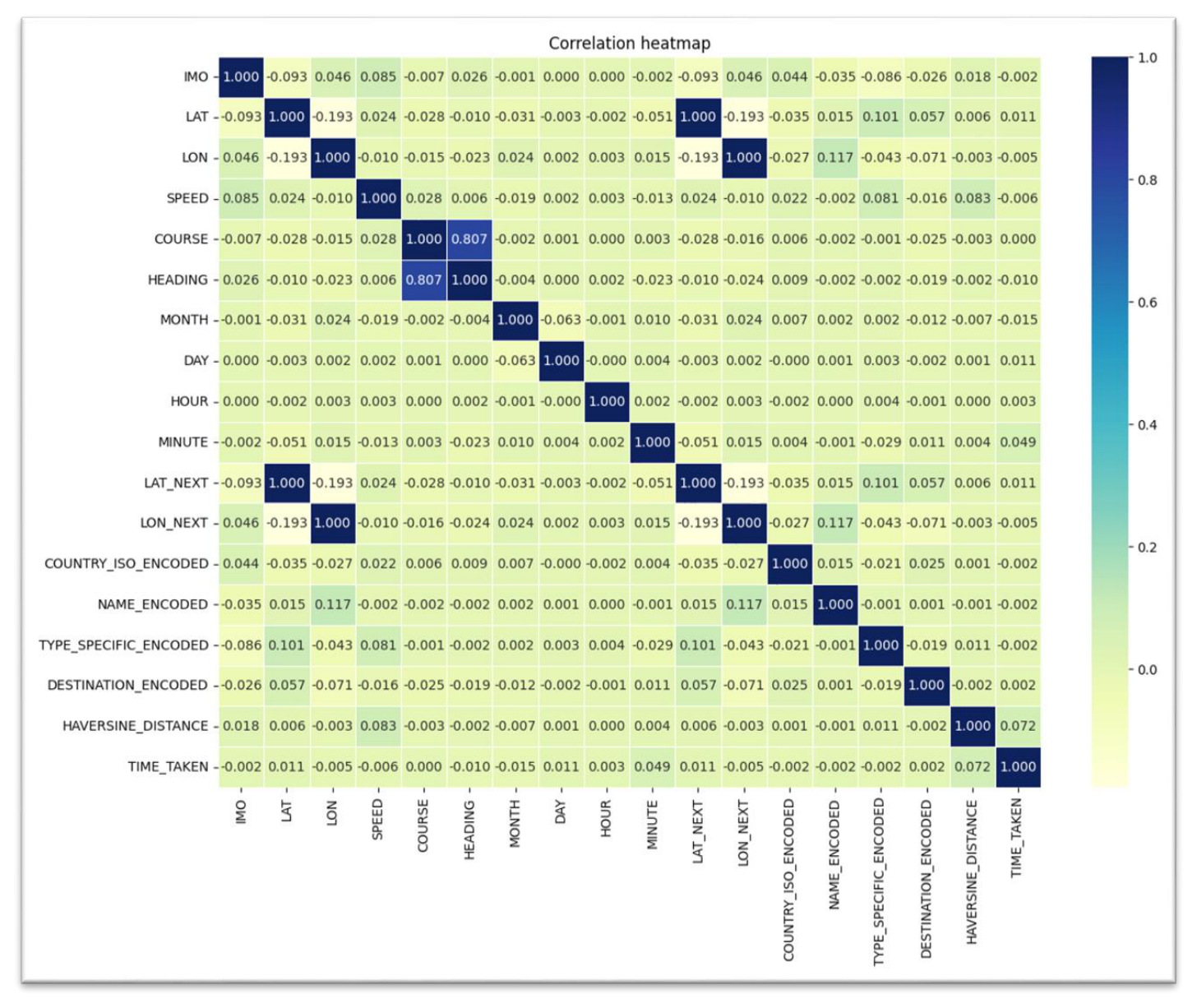}
  \caption{Correlation matrix}
  \label{fig:corr}
\end{figure}

The correlation graph (Figure~\ref{fig:corr}) shows high positive correlation between \texttt{TIME\_TAKEN} (target variable) and:
\begin{itemize}
  \item \texttt{HAVERSINE\_DISTANCE}
  \item \texttt{MINUTE}
  \item \texttt{LAT\_NEXT}
  \item \texttt{LAT}
  \item \texttt{DAY}
\end{itemize}

\subsection{Additional Features}

To improve the features to extract more from the data available, the following features are derived. These are already shown in the correlation graph above.
\begin{itemize}
  \item \texttt{LAT\_NEXT} -- the next available vessel latitude position in the data.
  \item \texttt{LON\_NEXT} -- the next available vessel longitude position in the data.
  \item \texttt{HAVERSINE\_DISTANCE} is a calculated feature. It gives the spherical distance (in km) between 2 geo-positional points on earth $(\mathrm{lat}_1, \mathrm{lon}_1)$ and $(\mathrm{lat}_2, \mathrm{lon}_2)$.
\end{itemize}

\subsection{Simple Linear Regression}

To start with, we do a simple linear regression to understand the feature importance and decide on the features to drop-off from the dataset. The simple linear regression results in the metrics in Table~\ref{tab:lrmetrics} and the feature importance in Table~\ref{tab:importance}.

\begin{table}[htbp]
  \centering
  \caption{Model metrics of simple linear regression}
  \label{tab:lrmetrics}
  \begin{tabular}{lr}
    \toprule
    Metric & Value \\
    \midrule
    Mean Squared Error & 88362.3148 \\
    Mean Absolute Percent Error (MAPE) & 14.75\% \\
    \bottomrule
  \end{tabular}
\end{table}

\begin{table}[htbp]
  \centering
  \caption{Feature importance after simple linear regression}
  \label{tab:importance}
  \begin{tabular}{lr}
    \toprule
    Feature & Importance \\
    \midrule
    \texttt{LAT\_NEXT} & 6.032160 \\
    \texttt{HAVERSINE\_DISTANCE} & 0.882499 \\
    \texttt{MINUTE} & 0.861984 \\
    \texttt{LON\_NEXT} & 0.470664 \\
    \texttt{DAY} & 0.342707 \\
    \texttt{HOUR} & 0.101297 \\
    \texttt{COURSE} & 0.064977 \\
    \texttt{DESTINATION\_ENCODED} & 0.000006 \\
    \texttt{NAME\_ENCODED} & $-0.000133$ \\
    \texttt{COUNTRY\_ISO\_ENCODED} & $-0.012753$ \\
    \texttt{TYPE\_SPECIFIC\_ENCODED} & $-0.025004$ \\
    \texttt{HEADING} & $-0.072077$ \\
    \texttt{LON} & $-0.485284$ \\
    \texttt{SPEED} & $-0.756354$ \\
    \texttt{MONTH} & $-1.560636$ \\
    \texttt{LAT} & $-5.873742$ \\
    \bottomrule
  \end{tabular}
\end{table}

\begin{figure}[htbp]
  \centering
  \includegraphics[width=0.85\textwidth]{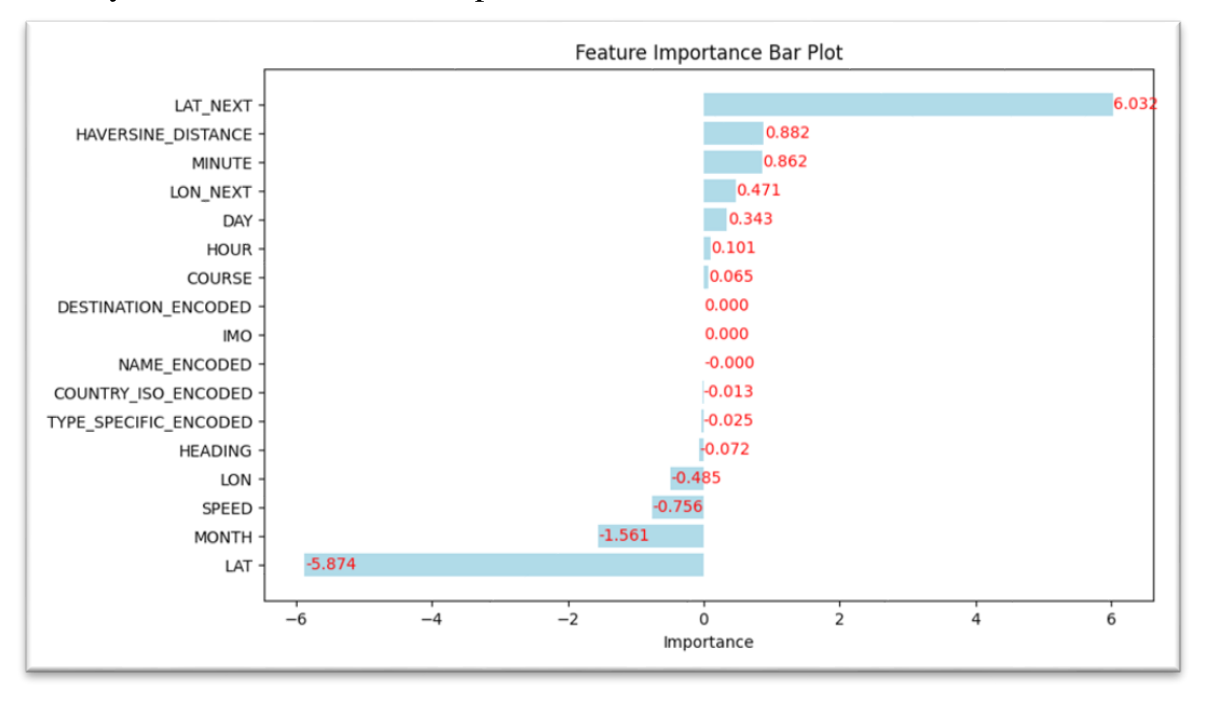}
  \caption{Feature importance}
  \label{fig:importance}
\end{figure}

Visually, Figure~\ref{fig:importance} shows the feature importance values.

We can conveniently drop 3 features which have no impact on the \texttt{TIME\_TAKEN} target variable. The features are \texttt{NAME\_ENCODED}, \texttt{DESTINATION\_ENCODED} and \texttt{IMO}.

\subsection{Principal Component Analysis}

\begin{figure}[htbp]
  \centering
  \includegraphics[width=0.95\textwidth]{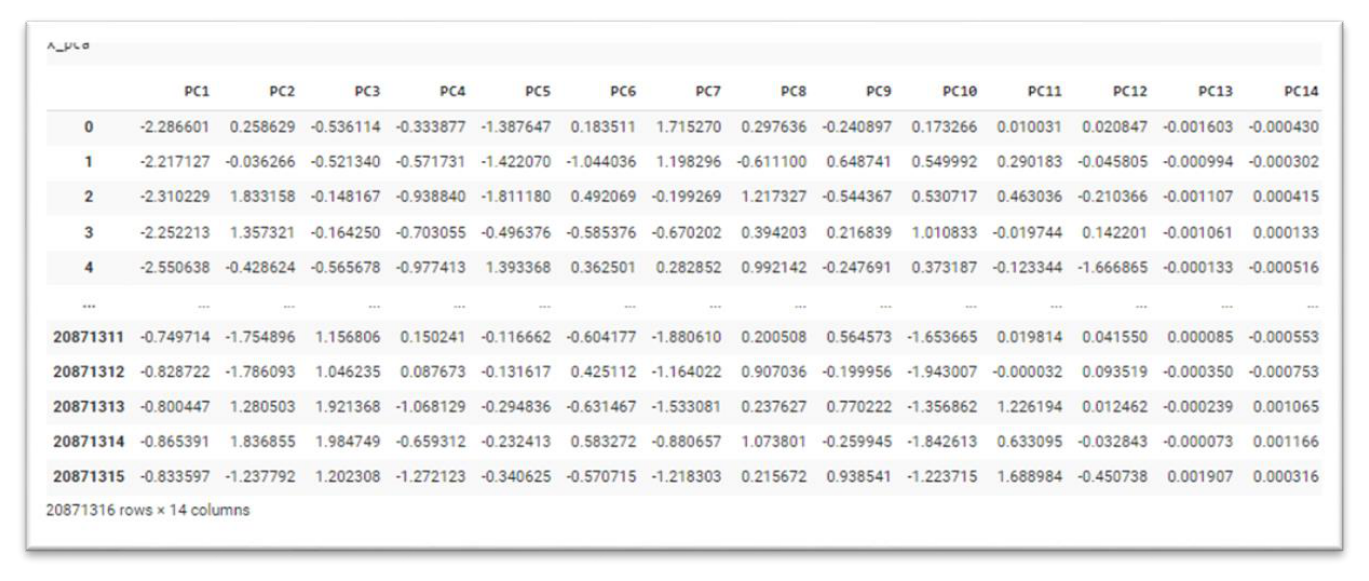}
  \caption{Principal component analysis}
  \label{fig:pca}
\end{figure}

Dimensionality reduction is done on the remaining 14 features using principal component analysis. The PCA results in 14 components. We use explained variance technique to drop a few components (Figure~\ref{fig:pca}).

Explained variance ratios with cumulative explained variance percent are shown in Table~\ref{tab:pca}.

\begin{table}[htbp]
  \centering
  \caption{Cumulative explained variance of PCA components}
  \label{tab:pca}
  \begin{tabular}{lrrr}
    \toprule
    Component & Explained Variance & Percent & Cumulative Percent \\
    \midrule
    PC1  & 0.172018078 & 17.20\% & 17.20\% \\
    PC2  & 0.130208212 & 13.02\% & 30.22\% \\
    PC3  & 0.115307599 & 11.53\% & 41.75\% \\
    PC4  & 0.079680702 & 7.97\%  & 49.72\% \\
    PC5  & 0.075772382 & 7.58\%  & 57.30\% \\
    PC6  & 0.072020458 & 7.20\%  & 64.50\% \\
    PC7  & 0.071503800 & 7.15\%  & 71.65\% \\
    PC8  & 0.071033642 & 7.10\%  & 78.75\% \\
    PC9  & 0.069195469 & 6.92\%  & 85.67\% \\
    PC10 & 0.066765730 & 6.68\%  & 92.35\% \\
    PC11 & 0.062773739 & 6.28\%  & 98.63\% \\
    PC12 & 0.013718362 & 1.37\%  & 100.00\% \\
    PC13 & 0.000001202 & 0.00\%  & 100.00\% \\
    PC14 & 0.000000625 & 0.00\%  & 100.00\% \\
    \bottomrule
  \end{tabular}
\end{table}

\begin{figure}[htbp]
  \centering
  \includegraphics[width=0.6\textwidth]{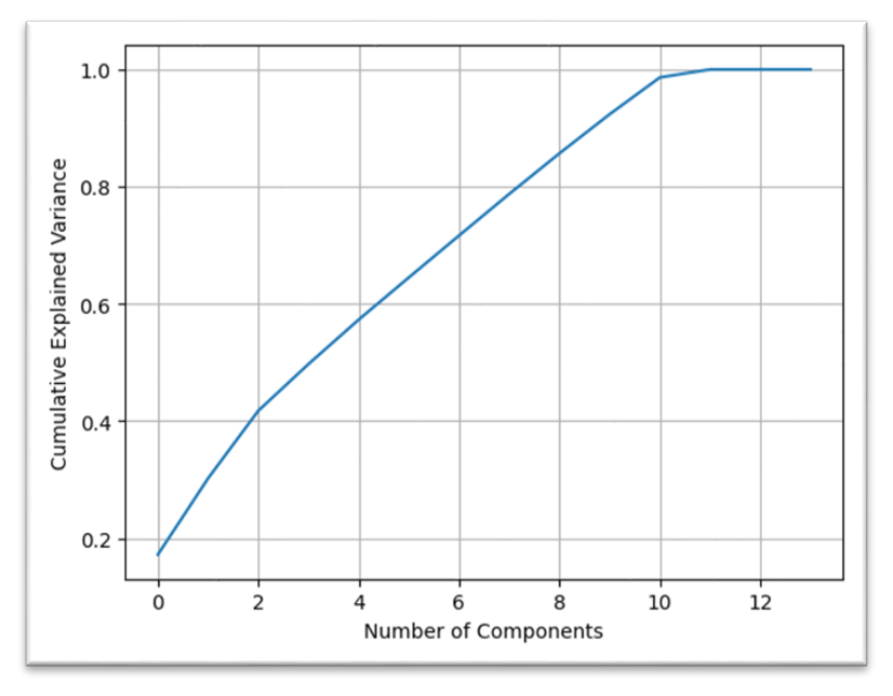}
  \caption{PCA explained variance}
  \label{fig:pcavar}
\end{figure}

From the distribution graph (Figure~\ref{fig:pcavar}), it can be seen that components PC13 and PC14 have no explained variance and can be dropped. The rest of the 12 components are used for 1D-CNN deep learning regression.

\subsection{1D-CNN Deep Learning Model}

\begin{figure}[htbp]
  \centering
  \includegraphics[width=0.55\textwidth]{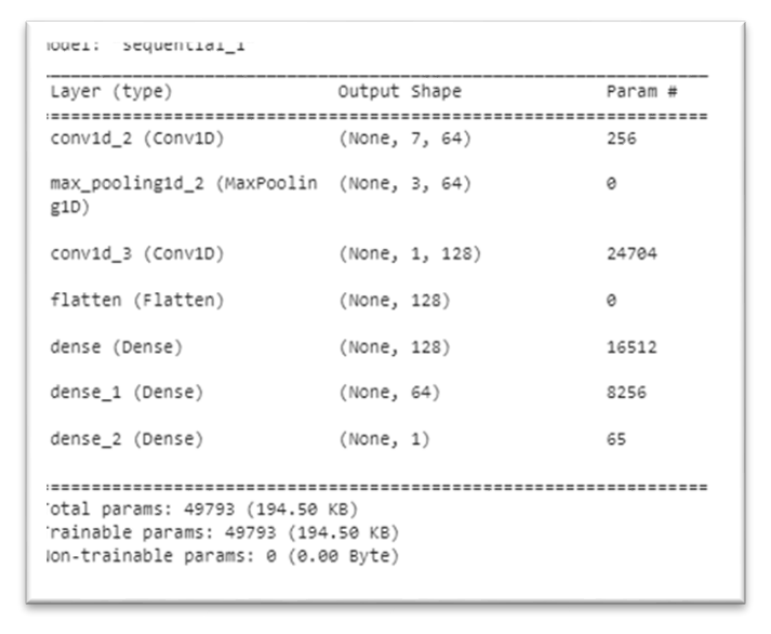}
  \caption{Neural network parameters}
  \label{fig:nnparams}
\end{figure}

\begin{figure}[htbp]
  \centering
  \includegraphics[width=0.45\textwidth]{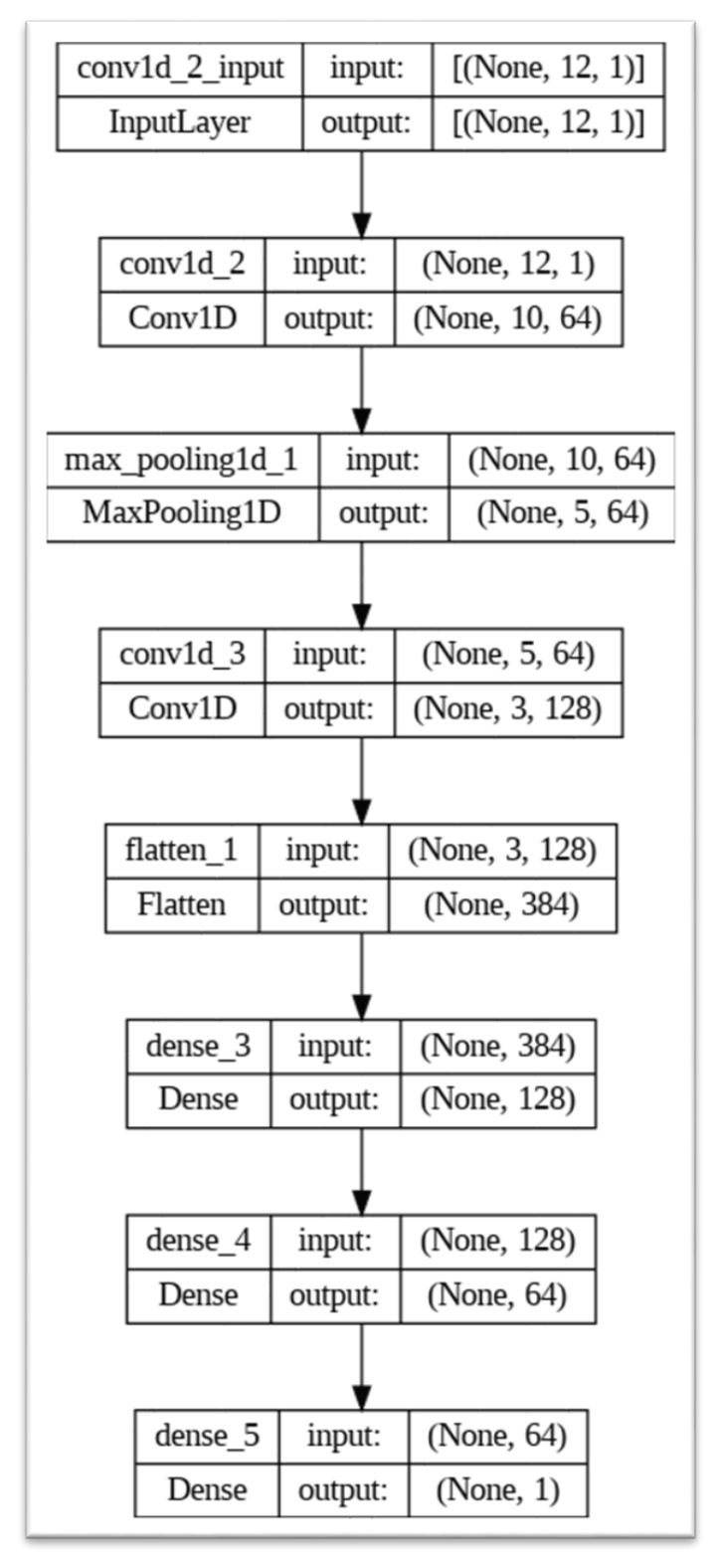}
  \caption{Neural network architecture}
  \label{fig:nnarch}
\end{figure}

We will use 2 hidden convolutional layers with ReLU as the activation function. The parameters of the model are given in Figure~\ref{fig:nnparams} and the neural network architecture is shown in Figure~\ref{fig:nnarch}.

\begin{figure}[htbp]
  \centering
  \includegraphics[width=0.95\textwidth]{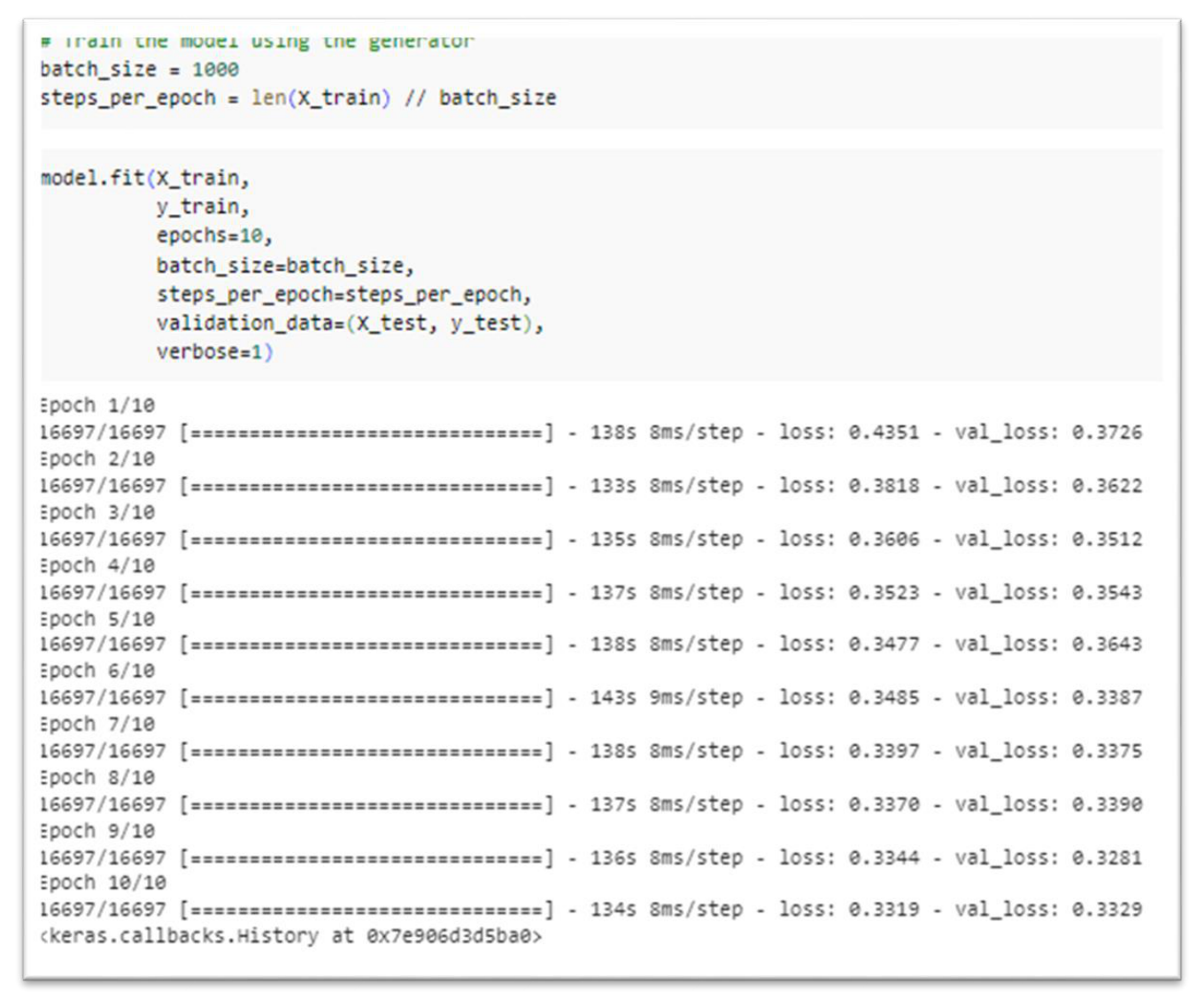}
  \caption{Model training}
  \label{fig:training}
\end{figure}

The dataset is scaled using standard scaler so that the neural network performs relatively better. With an A100 GPU processor, the training is done on 80\% of the dataset with a batch size of 1000. There are 10 epochs run and the validation loss reduces considerably with each epoch. The final validation loss is 0.3329 (Figure~\ref{fig:training}).

\begin{figure}[htbp]
  \centering
  \includegraphics[width=0.7\textwidth]{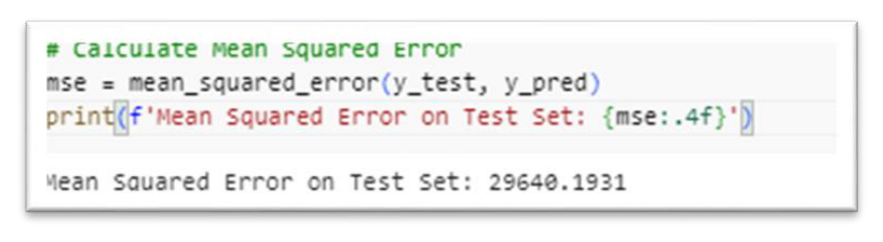}
  \caption{Model mean square error}
  \label{fig:mse}
\end{figure}

\begin{figure}[htbp]
  \centering
  \includegraphics[width=0.7\textwidth]{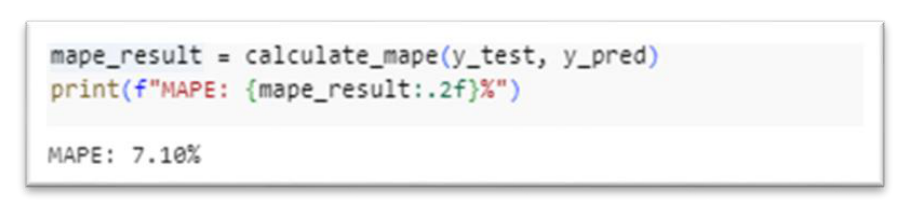}
  \caption{Model mean absolute percent error (MAPE)}
  \label{fig:mape}
\end{figure}

\begin{figure}[htbp]
  \centering
  \includegraphics[width=0.35\textwidth]{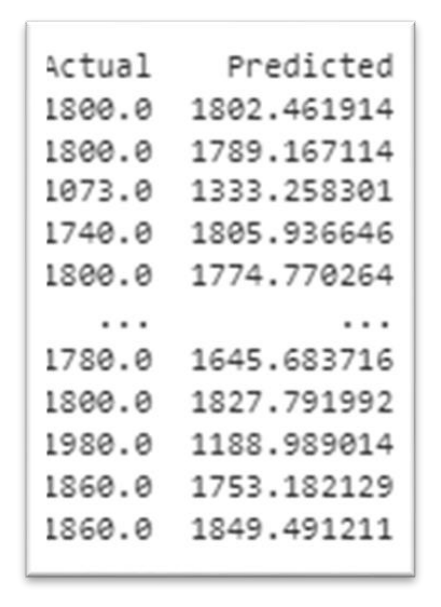}
  \caption{Actual and predicted values}
  \label{fig:actualpred}
\end{figure}

\begin{figure}[htbp]
  \centering
  \includegraphics[width=0.6\textwidth]{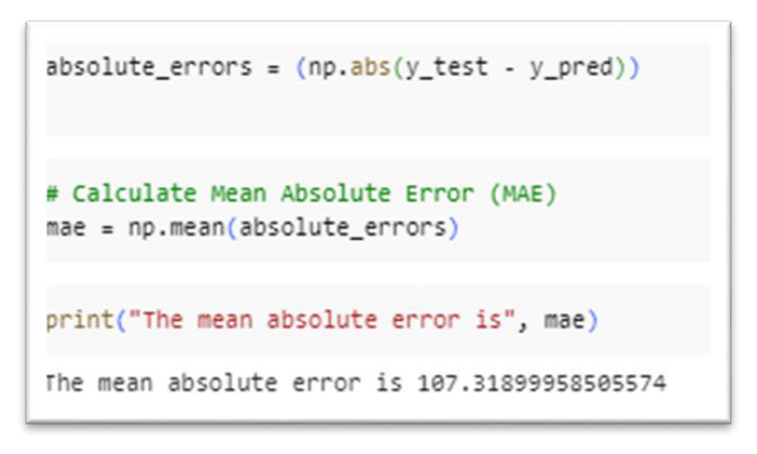}
  \caption{Model mean absolute error}
  \label{fig:mae}
\end{figure}

The final results of the 1D-CNN model are given in Table~\ref{tab:cnnresults}.

\begin{table}[htbp]
  \centering
  \caption{Results of 1D-CNN regression}
  \label{tab:cnnresults}
  \begin{tabular}{lr}
    \toprule
    Metric & Value \\
    \midrule
    Mean Squared Error & 29640.1911 \\
    Mean Absolute Percent Error (MAPE) & 7.10\% \\
    Mean Absolute Error (MAE) & 107.32 \\
    \bottomrule
  \end{tabular}
\end{table}

\subsection{Conclusion}

We have been successfully able to collect a year worth of transit data for 13000+ vessels and analyze using 22+ million datapoints. The 1D-CNN model shows good accuracy of 7.10\% MAPE and this model can be used for building a wrapper application to predictions from point A to point B.

To conclude, the research demonstrates the potential for using advanced machine learning techniques to analyze large amounts of transit data and make accurate predictions.

\section{Discussion}

\subsection{Discussion of Results}

The features listed in Table~\ref{tab:inputfeatures} have been used to build the model with a 7\% MAPE.

\begin{table}[htbp]
  \centering
  \caption{Input features of the model}
  \label{tab:inputfeatures}
  \begin{tabular}{l}
    \toprule
    Features used \\
    \midrule
    \texttt{LAT\_NEXT} \\
    \texttt{HAVERSINE\_DISTANCE} \\
    \texttt{MINUTE} \\
    \texttt{LON\_NEXT} \\
    \texttt{DAY} \\
    \texttt{HOUR} \\
    \texttt{COUNTRY\_ISO\_ENCODED} \\
    \texttt{TYPE\_SPECIFIC\_ENCODED} \\
    \texttt{HEADING} \\
    \texttt{LON} \\
    \texttt{SPEED} \\
    \texttt{MONTH} \\
    \texttt{LAT} \\
    \bottomrule
  \end{tabular}
\end{table}

The \texttt{TIME\_TAKEN} (in seconds) is the target variable. The \texttt{HAVERSINE\_DISTANCE} is a derived field. \texttt{LON\_NEXT} and \texttt{LAT\_NEXT} is the desired next latitude and longitude of the path of the vessel.

With the trained model, we can supply the features listed in Table~\ref{tab:appfeatures} to get the time taken from point A and point B. The wrapper application has to decide the path to be taken and the model predicts the time for the travel.

\begin{table}[htbp]
  \centering
  \caption{Features required in end application to use the model}
  \label{tab:appfeatures}
  \begin{tabular}{ll}
    \toprule
    Feature required & Description \\
    \midrule
    \texttt{LATITUDE} & Current latitude at point A \\
    \texttt{LONGITUDE} & Current longitude at point A \\
    \texttt{DATE-TIME} & Current datetime in UTC \\
    \texttt{LAT\_NEXT} & Planned latitude of point B \\
    \texttt{LON\_NEXT} & Planned longitude of point B \\
    \texttt{COUNTRY\_ISO} & Vessel's country ISO \\
    \texttt{TYPE\_SPECIFIC} & Subtype of CARGO vessel \\
    \texttt{HEADING} & Planned orientation of vessel \\
    \texttt{SPEED} & Planned average speed of vessel in knots \\
    \bottomrule
  \end{tabular}
\end{table}

The calling application should decide the path to be taken with the other features and the model determines the time taken at that step. A long journey should be split into multiple parts and each part time taken should be determined by the model. The overall time taken by the journey is the sum total of all the individual part time taken.

\subsection{Improvements Suggested}

We have built a basic model with huge data and good features. Some improvements of the model are possible if we consider the following:
\begin{itemize}
  \item Load on the vessel -- the total tonnage that is being shipped by the vessel including the container weight and ship weight.
  \item Traffic at the point of snapshot time -- the other ships in the radius of 50 km at the snapshot point latitude, longitude.
  \item Predicted/actual weather conditions at the latitude, longitude at the snapshot time. The actual weather conditions can be used during training and the predicted weather can be added as feature to the future data for predictions.
\end{itemize}

These features may provide some marginal improvements. The constructed model already has a good time prediction with plus/minus 7\% of the actual time taken. This is good enough to predict the final time of travel of a long haul journey of a vessel.

\section{Summary, Implications, and Recommendations}

\subsection{Summary}

To summarize, the following activities were undertaken to fulfil the objectives of the research. The objective of the research was to predict accurately the time taken between 2 points on a sea route to consolidate and find the total time taken for a cargo vessel to travel between two ports. This travel time is crucial for supply chain activities as there is a complex network of raw materials and finished goods flowing across the world for manufacturing businesses.

Activities:
\begin{enumerate}
  \item Collected data for 13000+ vessels with country and vessel name.
  \item Collected datapoints for transits of the 13000+ vessels during complete year 2023. There were 23 million data points collected with geo-positions, speed, timestamp and destination. There are 15 features in the collected data from API. The data provider used was datalastic.com.
  \item The data was cleaned to remove outliers and missing values.
  \item The categorical values were encoded and made ready for the model.
  \item Additional features were added according to the purpose of the model and business knowledge. The number of features now is 17.
  \item Correlation study of the features was done.
  \item Feature importance study was done using linear regression and 3 features with no importance were removed. The number of features ready for the model is 14.
  \item With principal component analysis, 12 features remain and are useful.
  \item 1D-CNN was done with 12 features and \texttt{TIME\_TAKEN} as target feature.
  \item The regression yielded good results of 7\% MAPE.
\end{enumerate}

\subsection{Implications}

The 1D-CNN model is a deep learning model that is particularly useful for extracting correlations between data features. When applied to vessel travel time prediction, the 1D-CNN model can help identify patterns in the data that may be difficult to detect using other methods. This can lead to more accurate predictions of travel time, which can be useful for a variety of applications, such as optimizing shipping routes, improving logistics planning, and reducing fuel consumption.

\subsection{Recommendations for Future Research}

Any new features can be added and correlated to the target variable. The variables can be studied further. Also, more data covering 3 years could be collected to improve the model.

Additional features possible are weather temperature, atmospheric pressure, vessel traffic and vessel tonnage. Anchor time of the vessel could be added in the final travel time to finalize the arrival time.

The 1D-CNN model can be combined with other deep learning models, such as LSTM and attention networks, to further improve the accuracy of travel time predictions.

\subsection{Conclusion}

The vessel travel time prediction was successfully done and a prediction with mean absolute percentage error of 7\% was obtained. The 1D-CNN model was implemented for the regression problem and it converged faster with the large amount of data used.

\section*{Acknowledgements}

This project would not have been possible without the support and help of numerous people. Firstly, I would like to thank my mentor and inspiration Dr.~Mario Silic who has been reviewing and supporting throughout my work. Next, my family has been very co-operative, allowing me to delve into this work day and night. They have been providing me necessary strength as well. For the data gathering part, the support team at datalastic.com have been my go-to people to help gather data and provide support to issues in the API for data gathering. To everyone else who have been helping me realize the importance of problem solving, research and critical thinking capabilities -- thank you. This would not have been possible without any of you.

\appendix

\section{Data Collection -- API and Code}

\subsection{Getting List of Vessels by IMO}

The following API is used to gather vessel data from the data provider datalastic.com. The API is subscription based and costs EUR~649 per month.

\medskip
\noindent API: \url{https://api.datalastic.com/api/v0/report}

\begin{lstlisting}
payload_json = {"api-key": API_KEY,
                "Report_type": "vessel_list"}
\end{lstlisting}

The API call returns a zip file with all the vessels tracked by the system. The columns provided in the CSV file are described in Table~\ref{tab:apivessels}.

\begin{table}[htbp]
  \centering
  \caption{API output for vessels list}
  \label{tab:apivessels}
  \begin{tabular}{lp{0.6\textwidth}}
    \toprule
    Column & Description \\
    \midrule
    \texttt{uuid} & Universally unique identifier of the vessel \\
    \texttt{mmsi} & Maritime Mobile Service Identity 9 digits number for a ship identification \\
    \texttt{imo} & International Maritime Organization 7 digits number for a ship identification \\
    \texttt{eni} & European Number of Identification \\
    \texttt{name} & Vessel name \\
    \texttt{name\_ais} & Vessel name \\
    \texttt{country} & Country of origin \\
    \texttt{callsign} & Vessel's number for vessel identification \\
    \texttt{vessel\_type} & Vessel type \\
    \texttt{vessel\_type\_specific} & Subtype of vessel \\
    \texttt{gross\_tonnage} & Gross Registered Tonnage, is a ship's total internal volume expressed in ``register tons'', each of which is equal to 100 cubic feet (2.83 m$^3$) \\
    \texttt{deadweight} & A measure of how much weight a ship can carry in tonnage \\
    \texttt{teu} & Container ship capacity is measured in twenty-foot equivalent units \\
    \texttt{length} & Length of the vessel \\
    \texttt{breadth} & Breadth of the vessel \\
    \texttt{home\_port} & Vessel's home port \\
    \texttt{year\_build} & Year ship was commissioned \\
    \texttt{status} & Ship active status \\
    \bottomrule
  \end{tabular}
\end{table}

\subsection{Getting Travel Geo-coordinates for One Year}

For a sample of 13000 IMOs from the vessel list, ship type of CARGO is filtered and all the activity of the vessel in the year 2023 is gathered with geo-positional data.

\medskip
\noindent API: \url{http://api.datalastic.com/api/v0/vessel_history}

\begin{lstlisting}
payload_json = {"api-key": API_KEY,
                "IMO": IMO,
                "FROM": FROM_DATE,
                "TO": TO_DATE}
\end{lstlisting}

For a hit on the API, there can be only 30 days of data retrieved. So, the API is hit for 12 times for each vessel IMO. The code for the construction of the tracking data frame is given below.

\begin{lstlisting}
import requests
import json
import pandas as pd

def get_data_by_imo(imo):
    imo = str(imo)
    months = [["2023-01-01", "2023-01-31"],
              ["2023-02-01", "2023-02-28"],
              ["2023-03-01", "2023-03-31"],
              ["2023-04-01", "2023-04-30"],
              ["2023-05-01", "2023-05-31"],
              ["2023-06-01", "2023-06-30"],
              ["2023-07-01", "2023-07-31"],
              ["2023-08-01", "2023-08-31"],
              ["2023-09-01", "2023-09-30"],
              ["2023-10-01", "2023-10-31"],
              ["2023-11-01", "2023-11-30"],
              ["2023-12-01", "2023-12-31"]]
    final_list = []
    for m in months:
        print(imo, m[0], m[1])
        from_date = m[0]
        to_date = m[1]
        # Construct the API URL with the given IMO number and date range
        url = "http://api.datalastic.com/api/v0/vessel_history?api-key=" + API_KEY \
              + "&imo=" + imo + "&from=" + from_date + "&to=" + to_date
        # Make a POST request to the API
        r = requests.post(url=url)
        results = r.text
        # Parse the API response as JSON
        results_json = json.loads(results)
        results_data = results_json["data"]
        # Extract relevant data from the API response
        country_iso = results_data["country_iso"]
        eni = results_data["eni"]
        imo = results_data["imo"]
        mmsi = results_data["mmsi"]
        name = results_data["name"]
        ship_type = results_data["type"]
        type_specific = results_data["type_specific"]
        uuid = results_data["uuid"]
        for p in results_data["positions"]:
            lat = p["lat"]
            lon = p["lon"]
            speed = p["speed"]
            course = p["course"]
            heading = p["heading"]
            destination = p["destination"]
            last_position_epoch = p["last_position_epoch"]
            last_position_UTC = p["last_position_UTC"]
            # Append the extracted data to the final list
            final_list.append([country_iso, eni, imo, mmsi, name,
                               ship_type, type_specific, uuid,
                               lat, lon, speed, course, heading,
                               destination, last_position_epoch,
                               last_position_UTC])
    # Create a pandas DataFrame from the final list
    final_df = pd.DataFrame(final_list, columns=["COUNTRY_ISO", "ENI", "IMO",
                                                 "MMSI", "NAME", "SHIP_TYPE",
                                                 "TYPE_SPECIFIC", "UUID",
                                                 "LAT", "LON", "SPEED",
                                                 "COURSE", "HEADING",
                                                 "DESTINATION",
                                                 "LAST_POSITION_EPOCH",
                                                 "LAST_POSITION_UTC"])
    return final_df
\end{lstlisting}

The constructed data frame has the structure described in Table~\ref{tab:dataframe}.

\begin{table}[htbp]
  \centering
  \caption{Data collected for the model}
  \label{tab:dataframe}
  \begin{tabular}{lp{0.55\textwidth}}
    \toprule
    Column & Description \\
    \midrule
    \texttt{COUNTRY\_ISO} & The ISO country code of the vessel's current location \\
    \texttt{ENI} & The European Vessel Identification Number \\
    \texttt{IMO} & The International Maritime Organization number, a unique identifier for ships \\
    \texttt{MMSI} & The Maritime Mobile Service Identity, a unique identifier for maritime communication \\
    \texttt{NAME} & The name of the vessel \\
    \texttt{SHIP\_TYPE} & The general type or category of the vessel \\
    \texttt{TYPE\_SPECIFIC} & Additional specific information about the vessel's type \\
    \texttt{UUID} & The Universally Unique Identifier, a unique identifier for the vessel \\
    \texttt{LAT} & The latitude coordinate of the vessel's position \\
    \texttt{LON} & The longitude coordinate of the vessel's position \\
    \texttt{SPEED} & The speed of the vessel in knots \\
    \texttt{COURSE} & The course or direction of the vessel in degrees \\
    \texttt{HEADING} & The heading or orientation of the vessel in degrees \\
    \texttt{DESTINATION} & The intended destination of the vessel \\
    \texttt{LAST\_POSITION\_EPOCH} & The timestamp of the vessel's last known position in epoch format \\
    \texttt{LAST\_POSITION\_UTC} & The timestamp of the vessel's last known position in UTC format \\
    \bottomrule
  \end{tabular}
\end{table}

\section{Data Pre-processing -- Code}

\subsection{Feature Engineering}

Firstly, only records with speed $>1$ knot were taken. This removed the records which showed that the vessel was stationary and not moving.

\begin{lstlisting}
final_df = pd.read_csv(file_path)
final = final[final.SPEED > 1]
final = final.drop(["MMSI", "SHIP_TYPE", "UUID", "ENI"], axis=1)
\end{lstlisting}

Since \texttt{SHIP\_TYPE} is all CARGO, it is dropped. \texttt{MMSI}, \texttt{UUID} and \texttt{ENI} are unique identifiers like \texttt{IMO}. Hence, they are also dropped, and \texttt{IMO} is retained. The following new features are created.

\begin{lstlisting}
### Add columns for the timestamp
final_df['YEAR'] = final_df['LAST_POSITION_UTC'].dt.year
final_df['MONTH'] = final_df['LAST_POSITION_UTC'].dt.month
final_df['DAY'] = final_df['LAST_POSITION_UTC'].dt.day
final_df['HOUR'] = final_df['LAST_POSITION_UTC'].dt.hour
final_df['MINUTE'] = final_df['LAST_POSITION_UTC'].dt.minute
\end{lstlisting}

The latitude and longitude change are gathered by taking leading values for each column.

\begin{lstlisting}
## Create a lead column for each IMO and DESTINATION
final_df["LAT_NEXT"] = final_df.groupby(['IMO', 'DESTINATION'])['LAT'].shift(1)
final_df["LON_NEXT"] = final_df.groupby(['IMO', 'DESTINATION'])['LON'].shift(1)
final_df["LAST_POSITION_EPOCH_NEXT"] = final_df.groupby(
    ['IMO', 'DESTINATION'])['LAST_POSITION_EPOCH'].shift(-1)
final_df["LAST_POSITION_UTC_NEXT"] = final_df.groupby(
    ['IMO', 'DESTINATION'])["LAST_POSITION_UTC"].shift(-1)
final_df = final_df[~final_df["LAST_POSITION_EPOCH_NEXT"].isna()]
\end{lstlisting}

Next, time taken (in seconds) is calculated. This is the target variable for the analysis.

\begin{lstlisting}
final_df["TIME_TAKEN"] = final_df["LAST_POSITION_EPOCH_NEXT"] \
                         - final_df["LAST_POSITION_EPOCH"]
\end{lstlisting}

To avoid outliers, the time taken $\leq 3000$ is only taken for modelling.

\begin{lstlisting}
final_df = final_df[final_df['TIME_TAKEN'] <= 3000]
\end{lstlisting}

Label encoding is done on the categorical values.

\begin{lstlisting}
label_encoder = LabelEncoder()
for column in ["COUNTRY_ISO", "NAME", "TYPE_SPECIFIC", "DESTINATION"]:
    print("Encoding column - ", column)
    encoded_column = column + "_ENCODED"
    final_df[encoded_column] = label_encoder.fit_transform(final_df[column])
\end{lstlisting}

Since we are dealing with latitude/longitude data, Haversine distance is calculated between 2 GPS points.

\begin{lstlisting}
final_df["HAVERSINE_DISTANCE"] = final_df.apply(haversine_distance, axis=1)
\end{lstlisting}

All the null values are set to zero for features \texttt{COURSE} and \texttt{HEADING}.

\begin{lstlisting}
final_df.COURSE = final_df.COURSE.fillna(0)
final_df.HEADING = final_df.HEADING.fillna(0)
final_df = final_df[~final_df.TIME_TAKEN.isna()]
\end{lstlisting}

The Haversine distance calculation function is given below.

\begin{lstlisting}
def haversine_distance(df):
    """
    Calculate the Haversine distance between two points on the Earth
    given their latitude and longitude.

    Parameters:
    - lat1, lon1: Latitude and longitude of the first point (in degrees)
    - lat2, lon2: Latitude and longitude of the second point (in degrees)

    Returns:
    - Distance between the two points in kilometers
    """
    # Convert latitude and longitude from degrees to radians
    lat1 = df["LAT"]
    lon1 = df["LON"]
    lat2 = df["LAT_NEXT"]
    lon2 = df["LON_NEXT"]
    lat1, lon1, lat2, lon2 = map(radians, [lat1, lon1, lat2, lon2])
    # Haversine formula
    dlat = lat2 - lat1
    dlon = lon2 - lon1
    a = sin(dlat / 2) ** 2 + cos(lat1) * cos(lat2) * sin(dlon / 2) ** 2
    c = 2 * atan2(sqrt(a), sqrt(1 - a))
    # Radius of the Earth in kilometers (mean value)
    radius = 6371.0
    # Calculate the distance
    distance = radius * c
    return distance
\end{lstlisting}

\section{Predictive Modelling -- Code}

\subsection{Simple Linear Regression}

\begin{lstlisting}
# Split the data into training and testing sets
X_train, X_test, y_train, y_test = train_test_split(X, y, test_size=0.3,
                                                    random_state=42)
lr = LinearRegression()
lr.fit(X_train, y_train)
y_pred = lr.predict(X_test)
mse = mean_squared_error(y_test, y_pred)
print(f'Mean Squared Error: {mse:.4f}')
mape_result = calculate_mape(y_test, y_pred)
print(f"MAPE: {mape_result:.2f} %")
\end{lstlisting}

\subsection{Principal Component Analysis}

\begin{lstlisting}
# Perform Principal component analysis
# Standardize the data
X_scaler = StandardScaler()
y_scaler = StandardScaler()
X_standardized = X_scaler.fit_transform(X)
y_standardized = y_scaler.fit_transform(
    np.array(y).reshape(-1, 1)).reshape(y.shape)

# Apply PCA
pca = PCA()
principal_components = pca.fit_transform(X_standardized)

# Create a DataFrame with the principal components
columns = [f'PC{i+1}' for i in range(principal_components.shape[1])]
X_pca = pd.DataFrame(data=principal_components, columns=columns)

print("\nExplained Variance Ratios:")
print(pca.explained_variance_ratio_)

cumulative_explained_variance = np.cumsum(explained_variance_ratio)
plt.plot(cumulative_explained_variance)
plt.xlabel('Number of Components')
plt.ylabel('Cumulative Explained Variance')
plt.grid(True)
plt.show()

n_components_to_retain = 9  # Replace with your chosen number
pca = PCA(n_components=n_components_to_retain)
X_pca = pca.fit_transform(X)
\end{lstlisting}

\subsection{1D-CNN Deep Learning}

\begin{lstlisting}
# Split the data into training and testing sets
X_train, X_test, y_train, y_test = train_test_split(X_pca, y_standardized,
                                                    test_size=0.2,
                                                    random_state=42)
model = Sequential()
model.add(Conv1D(filters=64, kernel_size=3, activation='relu',
                 input_shape=(10, 1)))
model.add(MaxPooling1D(pool_size=2))
model.add(Conv1D(filters=128, kernel_size=3, activation='relu'))
model.add(Flatten())
model.add(Dense(128, activation='relu'))
model.add(Dense(64, activation='relu'))
model.add(Dense(1, activation='linear'))

# Compile the model
model.compile(optimizer='adam', loss='mean_squared_error')
model.summary()

from tensorflow.keras.utils import plot_model
plot_model(model, to_file='model.png', show_shapes=True,
           show_layer_names=True)

# Train the model using the generator
batch_size = 64
steps_per_epoch = len(X_train) // batch_size
model.fit(X_train,
          y_train,
          epochs=10,
          batch_size=batch_size,
          steps_per_epoch=steps_per_epoch,
          validation_data=(X_test, y_test),
          verbose=1)

# Evaluate the model on the test set
y_pred = model.predict(X_test)
\end{lstlisting}

\subsection{RMSE and MAPE}

\begin{lstlisting}
def calculate_mape(y_true, y_pred):
    """
    Calculate Mean Absolute Percentage Error (MAPE).

    Parameters:
    - y_true: Actual values
    - y_pred: Predicted values

    Returns:
    - MAPE value
    """
    # Make sure both arrays are numpy arrays
    y_true = np.array(y_true)
    y_pred = np.array(y_pred)
    # Avoid division by zero
    mask = y_true != 0
    # Calculate absolute percentage error for each observation
    ape = np.abs((y_true - y_pred) / y_true)[mask]
    # Calculate the mean percentage error
    mape = np.mean(ape) * 100
    return mape

# Invert the scaling for predictions
y_pred = y_scaler.inverse_transform(y_pred).reshape(-1)
y_test = y_scaler.inverse_transform(y_test.reshape(-1, 1)).reshape(-1)

# Calculate Mean Squared Error
mse = mean_squared_error(y_test, y_pred)
print(f'Mean Squared Error on Test Set: {mse:.4f}')
print(pd.DataFrame(zip(y_test, y_pred)))
mape_result = calculate_mape(y_test, y_pred)
print(f"MAPE: {mape_result:.2f} %")
\end{lstlisting}

\end{document}